%% file: main.tex
\documentclass{article} 
\usepackage[final]{colm2026_conference}

\usepackage{microtype}
\usepackage{hyperref}
\usepackage{url}
\usepackage{booktabs}

\usepackage{lineno}
\usepackage{graphicx}

\usepackage{newfloat}
\usepackage{amsmath}
\usepackage{enumitem}
\usepackage{booktabs}
\usepackage{listings}
\usepackage{wrapfig}
\usepackage{multirow}

\definecolor{darkblue}{rgb}{0, 0, 0.5}
\hypersetup{colorlinks=true, citecolor=darkblue, linkcolor=darkblue, urlcolor=darkblue}

\title{FoR-SALE: Frame of Reference-guided Spatial Adjustment in LLM-based Diffusion Editing}

\author{Tanawan Premsri \& Parisa Kordjamshidi   \\
Department of Computer Science and Engineering\\
Michigan State University\\
East Lansing, MI 48824, USA \\
\texttt{\{premsrit,kordjams\}@msu.edu} \\
}

\begin{document}

\ifcolmsubmission
\linenumbers
\fi

\maketitle

\begin{abstract}
Current text-to-image generation models, even state-of-the-art models, exhibit a significant performance gap when spatial expressions are described from non-camera perspectives. To address this limitation, we propose \textbf{F}rame \textbf{o}f \textbf{R}eference-guided \textbf{S}patial \textbf{A}djustment in \textbf{L}LM-based Diffusion \textbf{E}diting (FoR-SALE), an extension of the Self-correcting LLM-controlled Diffusion (SLD). 
FoR-SALE first evaluates the alignment between a given text and an initially generated image, and then refines the image based on the expressed \textit{FoR} in the spatial description.
It employs vision modules to extract the spatial configuration of the generated image and simultaneously maps the spatial expression to a corresponding camera perspective.
This unified perspective enables direct evaluation of alignment between language and vision. 
When misalignment is detected, the required editing operations are generated and applied. FoR-SALE introduces novel latent-space operations to adjust the facing direction and depth of generated images.
We evaluate FoR-SALE on three benchmarks designed to assess spatial understanding with FoR: FoR-LMD, FoREST, and FoREST-G, with the latter providing diagnostic subsets covering multi-object scenes, expanded relation types, and naturalized prompts to test the generality of our framework.
Our framework improves the performance of SOTA T2I models by up to 7.0 pp using only a single round of correction.
\end{abstract}

\section{Introduction}
\input{01_introduction}

\section{Related Work}
\input{02_related_work}

\section{Methodology}
\input{03_methodology}

\section{Experiments}\label{sec:experiment_setting}

\input{04_experiment}

\input{05_extra_experiment}
\section{Conclusion}
Given the limitations of current text-to-image (T2I) models in handling spatial relations across diverse frames of reference (FoR), we propose FoR-SALE—Frame of Reference-guided Spatial Adjustment in LLM-based Diffusion Editing—to address this challenge.
Our framework extends the Self-correcting LLM-controlled Diffusion approach by introducing three key components: a comprehensive Visual Perception Module, a dedicated FoR Interpreter, and two new latent editing actions.
FoR-SALE can be applied to various T2I models and effectively improves the spatial alignment of images initially generated by those models, increasing accuracy by up to 7.0 pp after one correction round and 13.1 pp after three rounds.
Using GPT-4o as the base generator, our method achieves SOTA performance on spatial expressions involving FoRs, particularly for intrinsic FoRs, which are especially challenging. 
These results demonstrate the robustness of our proposed framework for FoR-aware spatial reasoning.

\section*{Acknowledgments}
This project is partially supported by the Office of Naval Research (ONR) grant N00014-23-1-2417. Any opinions, findings, conclusions, or recommendations expressed in this material are those of the authors and do not necessarily reflect the views of the Office of Naval Research. We thank anonymous reviewers for their constructive feedback, which greatly helped us improve this manuscript.

\bibliography{colm2026_conference}
\bibliographystyle{colm2026_conference}

\appendix

\input{91_Implementation_Details}

\input{92_Additional_results}

\input{99_prompt}

\end{document}

%% file: 01_introduction.tex
\begin{wrapfigure}{r}{0.45\textwidth}
    \centering
    \vspace{-2mm}
    \includegraphics[width=\linewidth]{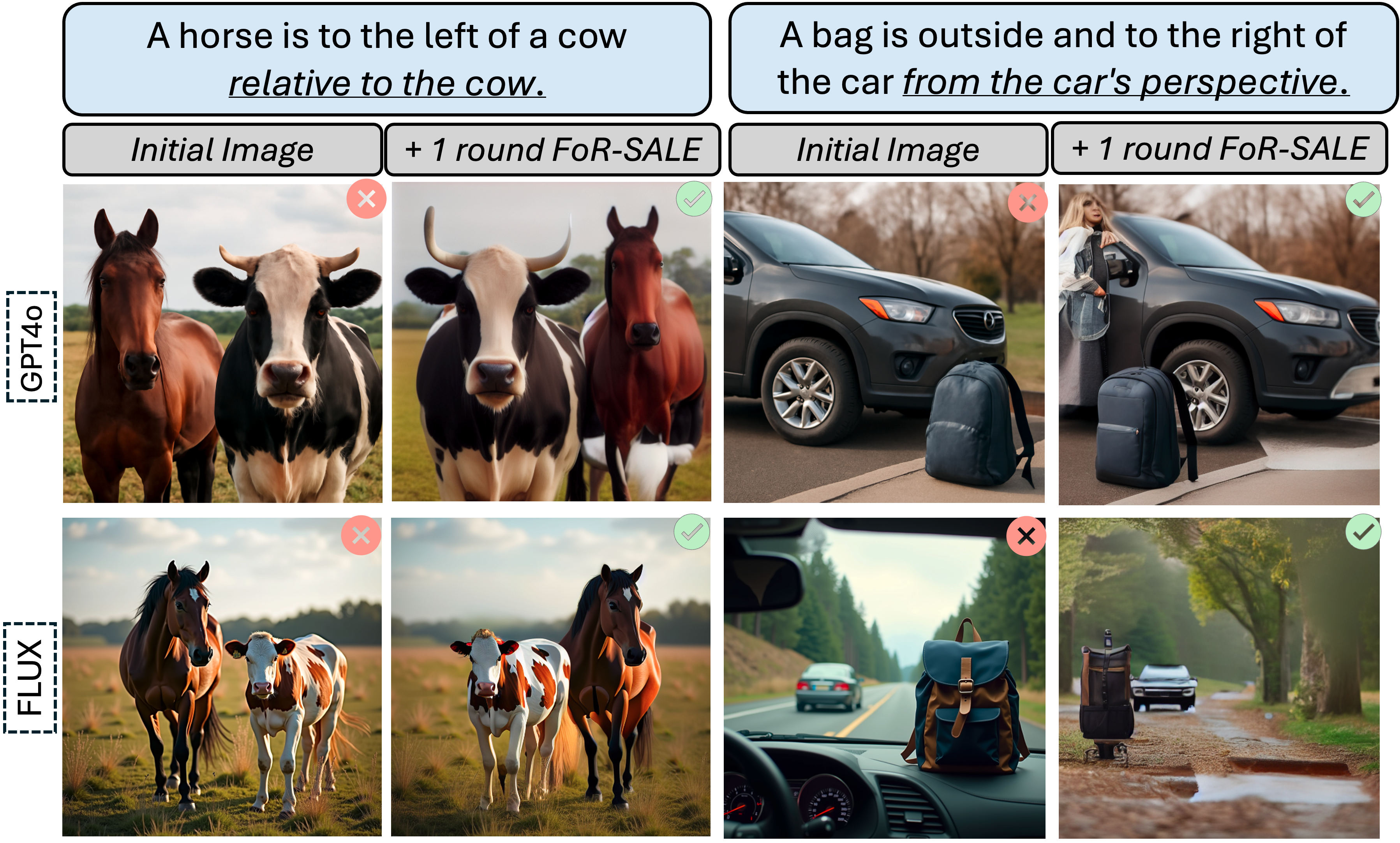}
    \caption{Examples of images generated by T2I models and the corresponding outputs after one round of correction with FoR-SALE.}
    \label{fig:example_failure}
     \vspace{-2mm}
\end{wrapfigure}

Spatial understanding refers to the ability to comprehend the locations of objects in space and is fundamental to human cognition and everyday tasks. 
One key component of spatial understanding is the Frame of Reference (FoR), which defines the perspective from which spatial relations are interpreted.
While extensively studied in cognitive linguistics~\citep{Mou2002-tp, Levinson_2003, TENBRINK2011704, cueBySeenAndHear}, FoRs have received limited attention in AI models, particularly within Multimodal Large Language Models (MLLMs)~\citep{VSR,SpatialVLM}.
Recent studies highlight shortcomings in FoR reasoning by MLLMs across multiple tasks, e.g., text-based Question Answering~\citep{FoREST}, Visual Question Answering~\citep{COMFORT}, and especially Text-to-Image (T2I) generation~\citep{GenSpace}. 
\citet{GenSpace} and~\citet{FoREST} show that diffusion models exhibit substantially lower spatial alignment when spatial descriptions are expressed from non-camera perspectives. 
As illustrated in Figure~\ref{fig:example_failure}, even SOTA T2I models, such as GPT-4o~\citep{GPT4o} and FLUX.1~\citep{blackforestlabs2024flux1dev}, struggle to accurately generate images that reflect spatial relations described from non-camera perspectives. 
To address this issue, we propose the \textbf{F}rame \textbf{o}f \textbf{R}eference-guided \textbf{S}patial \textbf{A}djustment in \textbf{L}LM-based Diffusion \textbf{E}diting (FoR-SALE) framework.
Our approach builds upon the Self-correcting LLM-controlled Diffusion (SLD) pipeline~\citep{Wu_SLD}, which uses LLMs as a core reasoning component to validate prompts and generate suggested layouts for editing images via latent-space operations in the diffusion model.
However, the original SLD framework does not account for FoR in spatial language, limiting its ability to handle spatial prompts grounded in non-camera perspectives. 
FoR-SALE extends this paradigm by explicitly modeling FoR and enabling spatial adjustments across diverse perspectives.

\begin{figure*}[t]
    \centering
    \includegraphics[width=0.7\linewidth]{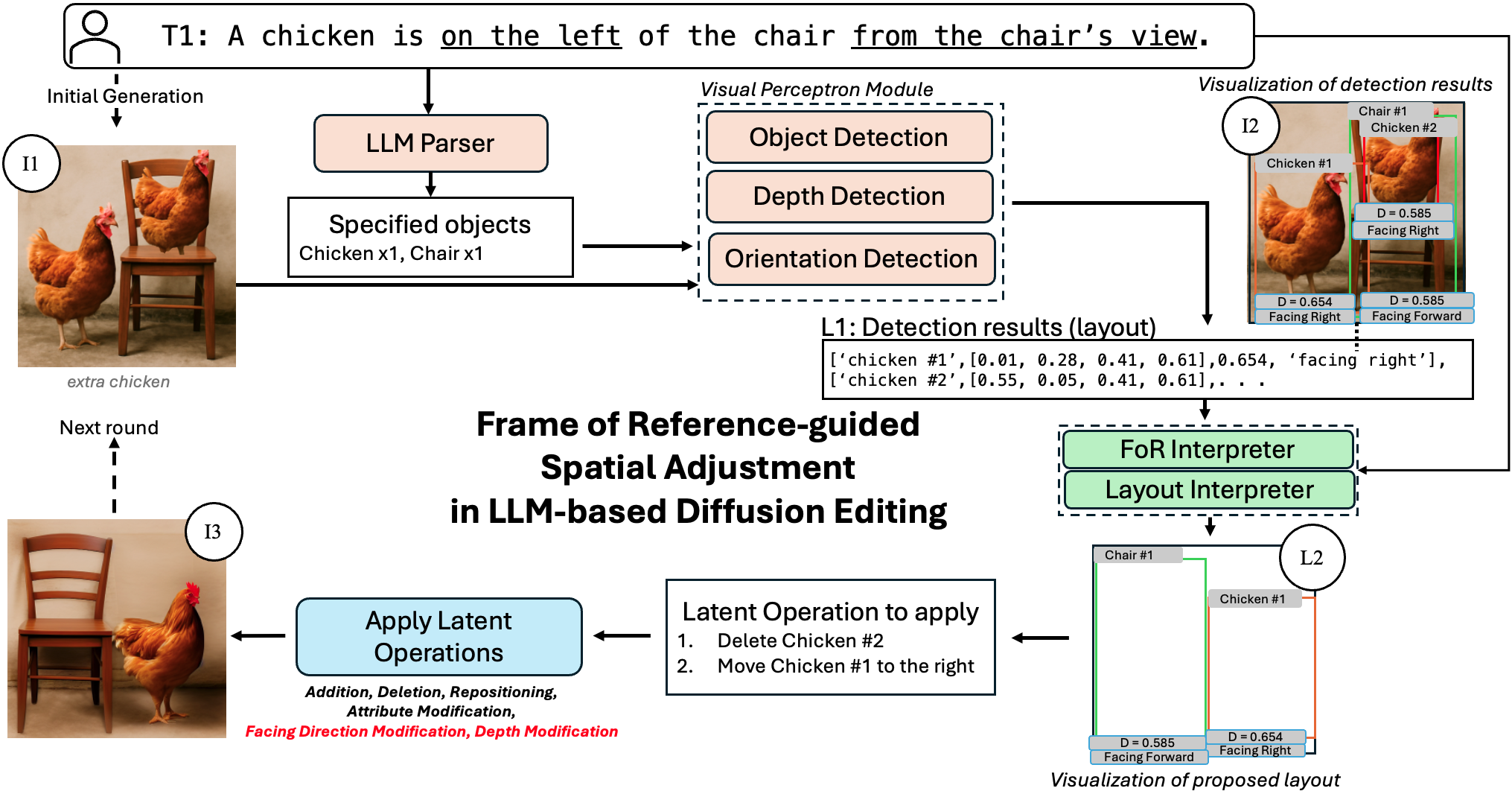}
    \caption{Overview of the FoR-SALE pipeline. It first extracts key objects from the input text using an LLM parser and layout information from the initial image using a Visual Perception Module. The extracted information is passed to the FoR Interpreter and Layout Interpreter to generate a revised layout. A sequence of latent operations is derived by comparing the initial and revised layouts and applied to synthesize an updated image. The resulting image can undergo additional refinement rounds if needed.} 
    \label{fig:pipeline}
\end{figure*}

Figure~\ref{fig:pipeline} illustrates the FoR-SALE pipeline.
It begins with standard T2I generation, where an input text ($T_1$) is passed to a T2I model to produce an image ($I_1$). 
Meanwhile, the LLM parser extracts key objects from the input text, which are then passed to the Visual Perception Module to obtain three visual properties: object location, orientation, and depth. 
These extracted visual properties ($I_2$) are converted into layout information ($L_1$). The input expression ($T_1$) and layout information ($L_1$) are then fed into the FoR Interpreter, which identifies the FoR and converts the expression into the camera perspective, a unified viewpoint.
Subsequently, the Layout Interpreter is employed
to generate a suggested layout ($L_2$) in textual form aligning with the updated expression. 
The suggested layout is then compared with the visual detection outputs ($L_1$) to identify mismatches, which are used to formulate self-correction operations, such as adjusting an object’s facing direction or depth.
These corrections are applied in the latent space during image synthesis using the Stable Diffusion model. 
Notably, these operations are generic and can be applied to other diffusion models. 
Finally, a new image is generated from the modified latent representation, ensuring consistency with the spatial configuration described in the input, particularly with respect to the specified FoR.
The resulting image ($I_3$) can undergo additional refinement rounds if needed.

{We evaluate the effectiveness of FoR-SALE on three benchmarks: FoR-LMD, a modification of LMD~\citep{LMD} that incorporates perspective cues; FoREST~\citep{FoREST}, a benchmark with textual inputs covering diverse FoR cases; and FoREST-G, a diagnostic extension of FoREST that evaluates generalization to multi-object scenes, expanded relation types, and naturalized prompts. 
FoR-SALE improves images generated by SOTA T2I models, including FLUX.1, Nano-Banana-Pro, and GPT-4o, by up to 7.0 pp in a single correction round and 13.1 pp in three rounds. Results on FoREST-G further demonstrate that FoR-SALE remains effective beyond the original two-object setting, particularly for intrinsic-FoR cases.}
Finally, we provide a thorough analysis of the limitations of both T2I models and LLMs in reasoning over layouts from different perspectives.
Our contributions\footnote{Code and data are available at \hyperlink{https://github.com/HLR/FoR-SALE}{https://github.com/HLR/FoR-SALE}.} can be summarized as follows:
\noindent\textbf{(1)} We propose the first self-correcting framework that explicitly incorporates frame of reference (FoR) into text-to-image generation. 
\noindent\textbf{(2)} We introduce novel editing operations within a self-correcting framework to handle FoR-aware image generation.
\noindent\textbf{(3)} We augment an existing benchmark to enable a more comprehensive evaluation of FoR understanding in text-to-image models. Our framework consistently improves FoR-aware spatial alignment, particularly under intrinsic FoR settings.

%% file: 02_related_work.tex
\textbf{Frame of Reference in MLLMs.}
Multiple benchmarks have been developed to evaluate the spatial understanding of MLLMs across various tasks~\citep{mattersim, mirzaee-etal-2021-spartqa, mirzaee-kordjamshidi-2022-transfer, shi2022stepgamenewbenchmarkrobust, vpeval}.
However, most of these benchmarks overlook FoR.
Only a few recent benchmarks explicitly address FoR-related reasoning~\citep{VSR, SpatialVLM, zhang2025spartund, Spatial457}.
For example, \citet{VSR} show that training a vision-language model with text that includes FoR information can improve visual question answering (VQA).
\citet{Spatial457} introduce a comprehensive benchmark for spatial VQA that includes FoR examples, although FoR is not its primary focus of evaluation.
Three recent studies focus more directly on evaluating FoR understanding in MLLMs.
First, \citet{COMFORT} assess FoR handling in VQA settings and reveal substantial limitations, especially when reasoning goes beyond the default camera-centric view.
Second, \citet{FoREST} investigate FoR reasoning in natural language prompts—both ambiguous and unambiguous—and find persistent failures in both question answering and layout generation when the perspective diverges from the camera view.
Third, \citet{GenSpace} evaluate T2I models and find that even SOTA models fail to preserve correct spatial relations when the context is not grounded in the camera perspective and includes 3D information such as orientation and distance.
In this work, we extend this line of research by providing a new evaluation of T2I models based on their alignment with FoR-grounded spatial expressions.
We also enhance LLM and T2I models' ability to comprehend varying FoR conditions.

\noindent\textbf{Spatial Alignment in T2I.}
Several studies have sought to improve the spatial alignment of T2I models with user input.
Early approaches introduced predefined spatial constraints—such as depth maps~\citep{ControlNet, FreeControl_2024}, object layouts~\citep{li2023gligen}, or attention maps~\citep{AttentionMapWang, pang2024attndreambooth}—to guide image generation. 
However, these often require manual configuration or model retraining to interpret the constraints.
With advances in LLMs' spatial reasoning, recent work has leveraged them to automatically generate spatial guidance.
For example, \citet{vpeval} use an LLM to generate layouts that guide diffusion models without additional training.
More recent methods employ MLLMs to control 3D spatial arrangements by generating feedback for reinforcement training of diffusion models~\citep{CoT-liz}, training a T2I model using compositional questions derived from the input~\citep{DreamSync}, or producing action plans for sequential editing~\citep{Wu_SLD, Grape}.
While these methods are promising, they do not address spatial FoR.
In contrast, we explicitly address this by extending the SLD framework~\citep{Wu_SLD} to support editing under diverse FoRs.

%% file: 03_methodology.tex
FoR-SALE follows the SLD framework~\citep{Wu_SLD}, which consists of two main components: (1) LLM-driven visual perception and (2) LLM-controlled Diffusion.
We further adapt these two components to enable finer-grained perception and layout interpretation for FoR recognition and image correction.
 
\subsection{LLM-driven Visual Perception Module}\label{sec:visual_detect_module}
The process begins with standard T2I generation, in which a textual input is used to create an image.
FoR-SALE then proceeds to extract key information from the spatial expression via an LLM Parser and from the generated image via a visual perception module.

\textbf{1) LLM Parser.} 
In this first step, we prompt an LLM with instructions and in-context examples to extract a list of key object mentions and their attributes from the input text, denoted as  $L$. 
For example, given the spatial expression \textit{A red chicken is on the left of a chair from the chair's view}, the output of the LLM is \textit{L = [(``chicken'', [``red'']), (``chair'', [None])]} where ``red'' is the attribute associated with the chicken, and ``None'' indicates that no specific attribute is mentioned for the chair. All prompt specifications are provided in Appendix~\ref{sec:LLM_prompt}.

\begin{figure*}[t]
    \centering
    \includegraphics[width=0.9\linewidth]{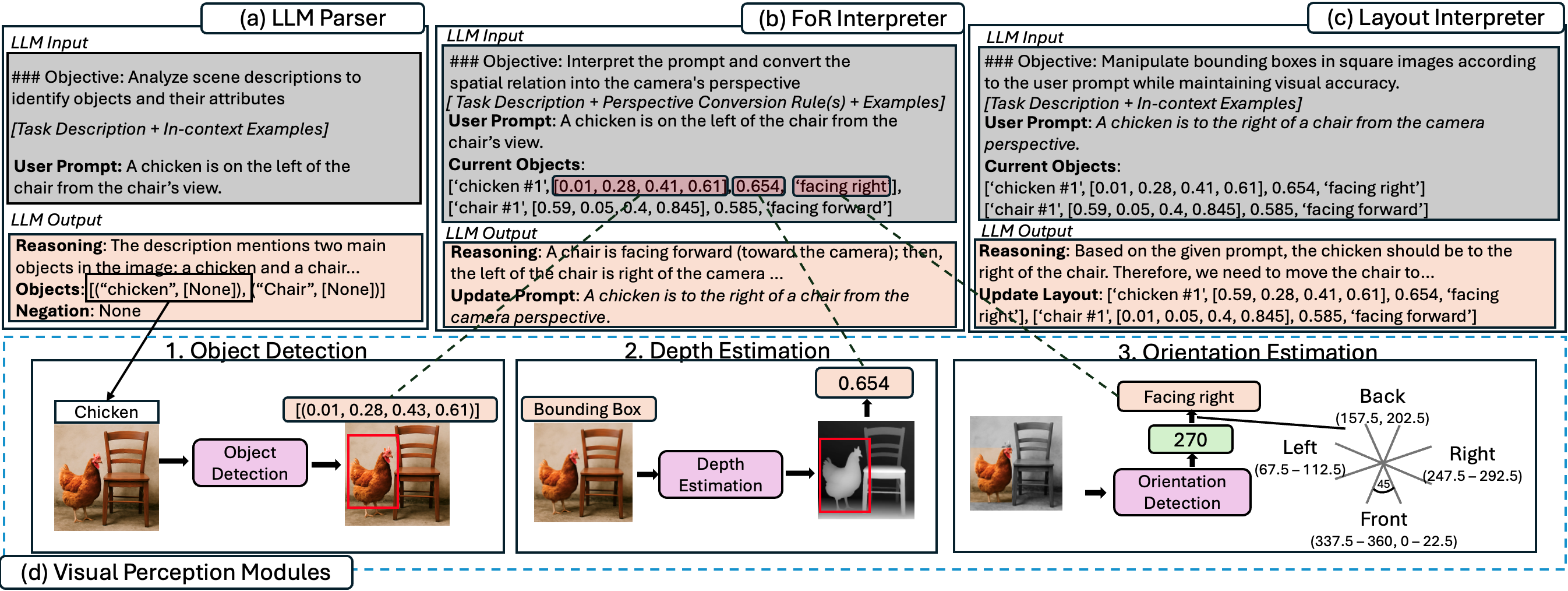}
    \caption{Example inputs and outputs from the LLM Parser, FoR Interpreter, Layout Interpreter, and Visual Perception Module.
The LLM Parser output guides the Visual Perception Module in extracting object-specific information, including bounding boxes, orientation, and depth.
This information is then passed to the FoR Interpreter, which converts the spatial expression to the camera perspective. The Layout Interpreter subsequently generates a suggested spatial layout based on the converted expression.}
    \label{fig:module_explain}
\end{figure*}

\textbf{2) Visual Perception Module.}\label{sec:visual_perception}
The obtained list $L$ is fed into the visual perception module with open-vocabulary object detection. In our FoR-SALE, we add new visual perception components to deal with FoR. These include depth estimation and orientation detection. 
Figure~\ref{fig:module_explain} (d) illustrates this module. 
The open-vocabulary object detector receives the information in $L$ in the following format 
``image of a/an [attribute] [object name]'' and outputs
bounding boxes, denoted as $B$. 
The outputs are represented in the following list format,
(\texttt{(attribute) (object name) (\#object ID)}, [$x$, $y$, $w$, $h$]) where ($x$, $y$) indicates the coordinates of the upper-left corner of the bounding box from 0.0 to 1.0, $w$ is its width, and $h$ is its height. The object ID is a serial number assigned uniquely to each detected object.
Next, the depth estimation model is used to predict the depth map, denoted as $D$. 
To extract object-specific depth values ($D_i$), a segmentation mask is obtained using the bounding boxes from $B$, and the average pixel depth within each region is computed using the following equation: $D_i = \frac{1}{ |R_i|}\sum_{j \in R_i} d_j  $, where $i$ is the index of the object, $R_i$ is the mask region of the object, and $d_j$ is the depth at pixel $j$. 
The value of $D_i$ ranges from 0 to 1.
Finally, an orientation detection model is applied to the object segmentation to obtain the object's orientation angle.
This angle is then converted into a facing direction, denoted as $f_i$. 
There are eight facing-direction categories: $orientation$=\{ForwardLeft, Left, BackwardLeft, Back, BackwardRight, Right, ForwardRight, Front\}.
Each category spans a 45-degree range, starting from 22.5° to 67.5° for ForwardLeft, and continuing in 45° intervals for the remaining orientation labels.
We collect this visual information for each object and obtain a new information list in a new format, denoted $V_L=$ \{(\texttt{(attribute) (object name) (\#object ID)}, [$x$, $y$, $w$, $h$], $D_i$, $f_i$)\}.
An example of representation is shown in Figure~\ref{fig:module_explain}.

\subsection{LLM-Controlled Diffusion}\label{sec:FoR_interpreter}
After obtaining visual information ($V_L$), two additional modules are employed to analyze and modify the image: LLM Interpreters and Image Correction.

\textbf{1) LLM Interpreters.} 
This module analyzes $V_L$ together with the input text $T$ and proposes a revised layout, denoted as $\tilde{V}_{L}$ in the same format.
The original SLD framework employs an LLM for layout interpretation.
However, in FoR-SALE, we incorporate an additional LLM, namely the FoR Interpreter. Figure~\ref{fig:module_explain} (b) and (c) illustrate these two LLMs.

\noindent\textbf{\textbullet \, FoR Interpreter.}
Based on findings that MLLMs demonstrate stronger performance when reasoning about spatial expressions described from a camera perspective~\citep{COMFORT, FoREST, GenSpace}, we hypothesize that converting spatial expressions to a camera perspective can enhance FoR reasoning. 
The FoR Interpreter takes the spatial text $T$ and visual information from the generated image $V_L$ as input, and outputs a rewritten spatial expression from the camera perspective, denoted as $T'$. 
If no spatial relation is present, the model returns the unchanged input.
We provide in-context information to guide the FoR Interpreter in performing this perspective conversion. 
In particular, we include spatial perspective conversion rules. A total of 32 rules are manually defined, one for each combination of the eight facing directions considered in the Visual Perception Module and four spatial relations (front, back, left, right).
For example, \textit{if the object is facing left, the left side of the object corresponds to the front direction from the camera perspective}. 
These rules cover directional spatial relations that are most strongly affected by FoR interpretation, as well as the eight possible facing directions. This set of spatial relations is based on qualitative directional relations, which are a closed set, making the formalization feasible~\citep{kordjamshidi-etal-2010-spatial}.
An example of the input and output is shown in Figure~\ref{fig:module_explain}(b), and all rules are included in Appendix~\ref{appendix:rules}. 
{To examine whether these rules can be automatically generated, we further evaluate their extensibility through an additional experiment that uses an LLM to generate rules automatically. The results and discussion are provided in RQ3 of Section~\ref{sec:experiment_results}.}

\noindent\textbf{\textbullet \, Layout Interpreter.}
After obtaining the spatial expression, $T'$, that follows the camera perspective, the second LLM uses $T'$ and $V_L$ as input to analyze the layout. 
The Layout Interpreter LLM is prompted with manually crafted in-context examples to analyze whether the current layout aligns with the provided $T'$. 
If misalignment is detected, the LLM is instructed to propose a revised layout $\tilde{V}_{L}$ that satisfies the spatial description. An example of the input and output is shown in Figure~\ref{fig:module_explain}(c).

\textbf{2) Image Correction.}
In this step, we compare the current layout $V_{L}$ with the proposed layout $\tilde{V}_{L}$ using an exact-matching process to detect misalignment.
If there is any misalignment between the two layouts, we create a sequence of editing operations to modify the image and align it with $\tilde{V}_{L}$.
The original SLD framework includes four editing operations: Addition, Deletion, Reposition, and Attribute Modification.
Our framework extends this set by introducing \textbf{two new operations for handling FoR}: Facing Direction Modification and Depth Modification.
Before applying any operation, backward diffusion~\citep{DiffusionModel} is performed on the initial image to obtain its latent representation, which serves as the basis for all subsequent editing actions.
After all editing actions are applied, Stable Diffusion is called to synthesize the final image.

\textbf{\textbullet \, Addition.} The operation first generates the object within the designated bounding box using the base Stable Diffusion model, after which its segment is extracted using SAM~\citep{SAM}. Next, a backward diffusion process is applied to the generated object segment to obtain an object-specific latent representation. This representation is then merged into the original image's latent space to complete the composition.

\textbf{\textbullet \, Deletion.} 
The process first segments the object using SAM within its bounding box. The latent representation of the segmented region is then removed and replaced with Gaussian noise, enabling reconstruction of the object region during the final diffusion step.

\textbf{\textbullet \, Reposition.} 
To preserve the object’s aspect ratio, this step begins by shifting and resizing the object from its original bounding box to the new bounding box.
After repositioning, SAM is used to segment the object. Then, a backward diffusion process is used to obtain the latent representation. This new representation is then integrated into the image latent space at the updated location.
To remove the original object, we replace the corresponding latent region segmented via SAM with Gaussian noise prior to the final diffusion step.

\textbf{\textbullet \, Attribute Modification.} 
The process begins by employing SAM to segment the object region within its bounding box.
An attribute modification diffusion model, e.g., DiffEdit~\citep{couairon2023diffedit}, is then called with a new prompt to modify the object’s attribute within the defined region.
After the attribute is edited, a backward diffusion process is performed to extract the corresponding latent representation.
This updated latent is then integrated into the image latent space to complete the modification.

\noindent\textbf{\textbullet \, Facing Direction Modification.} This new operation begins by using SAM to segment the object’s region. Then, DiffEdit is invoked with a prompt specifying the desired facing direction to generate an image of the object in that orientation.
Next, the base diffusion model is used to perform a backward diffusion process for obtaining the latent representation. Finally, this latent is integrated into the image latent space to complete the modification.

\noindent\textbf{\textbullet \, Depth Modification.} This new operation begins by synthesizing the new depth of the given object using the equation, 
\noindent$d_{j'} = min(1, \, max(0, \,d_{j} - D_{i} + D_{i'}))$,
where $d_j$, $d_{j'}$ denote the original and updated depth of pixel $j$, respectively.
$D_{i}$ represents the current average depth of object $i$ defined in Section~\ref{sec:visual_perception}, and $D_{i'}$ is the new target depth proposed by the Layout Interpreter.
Next, we shift and resize the synthesized depth map of this object to the target bounding box.
A diffusion model is then called with ControlNet~\citep{ControlNet} to generate an object with the specified depth.
After generating a new object, the segmentation and backward diffusion are performed to obtain the latent representation.
Finally, this latent representation is integrated into the image latent space to complete the modification.


%% file: 04_experiment.tex
\subsection{Datasets}

\noindent\textbf{FoR-LMD} is a synthetic dataset designed to assess several skills, including spatial understanding. 
We extend the LMD benchmark by adding perspective cues to the input template to incorporate FoR information. The modified prompt template is: \textit{($obj_1$) ($R_1$) ($ref_1$) and ($obj_2$) ($R_2$) ($ref_2$)}, where $ref_1$ and $ref_2$  specify the reference perspective, camera view (relative) or object-centric view (intrinsic). 
To focus on perspective-sensitive relations, we restrict $R_1$, $R_2$ to \{\text{left, right, front, back}\}.
This results in 500 samples with explicit perspective.

\noindent\textbf{FoREST}~\citep{FoREST} is a synthetic benchmark designed to evaluate FoR understanding with explicit FoR annotations (C-Split). We sample 500 spatial expressions to match the size of FoR-LMD. Each prompt explicitly specifies the spatial perspective and the facing direction of the reference object, which are not provided in FoR-LMD.

{
\noindent\textbf{FoREST-G} extends FoREST with three subsets: Complex, Relation-Augmented, and Language-Variation.
These subsets are designed to evaluate generalization beyond the original two-object setting and fixed prompt templates.
The Complex subset increases the number of objects and the complexity of spatial relations.
The Relation-Augmented subset adds vertical and diagonal relations, while the Language-Variation subset uses an LLM to rewrite the original spatial descriptions while preserving the target spatial constraints.
Further details on the construction of these subsets are provided in Appendix~\ref{appendix:diag_subset}.}


\subsection{Evaluation Method}\label{sec:evaluation_method}
We adopt the evaluation scheme proposed by~\citet{GenSpace}, which has been shown to align with human judgment. 
To evaluate generated images, we use the Visual Perception module (Section~\ref{sec:visual_detect_module}) to extract bounding boxes, depth, and object orientations.
We first verify that the number of detected objects matches the textual specification; in the evaluated benchmarks, exactly one instance of each object must be present. If this condition is not satisfied, the image is marked as incorrect.
We then compare detected orientations with annotated labels; any misalignment results in the image being marked as incorrect. 
Spatial relations are evaluated using the provided FoR annotations. When the FoR is non-camera-centric (intrinsic), spatial relations are converted to the camera perspective using the detected orientation of the reference object.
Finally, spatial correctness is assessed using predefined geometric specifications of spatial relations under the camera-perspective assumption~\citep{huang2023t2icompbench, vpeval, GenSpace}.
To validate our automatic evaluation, we conducted a human study on 200 randomly sampled images. The resulting Cohen’s kappa score of 0.791 indicates strong agreement between human judgments and the automatic evaluation. Details of the study are provided in Appendix~\ref{sec:human_alignmet_score}.

\subsection{Baseline Models}
For baseline comparison, we select seven T2I models: Stable Diffusion (SD) 1.5~\citep{SD1-5}, SD 2.1~\citep{SD1-5}, SD 3.5-Large~\citep{stabilityai2024sd35large}, GLIGEN~\citep{li2023gligen}, FLUX.1~\citep{blackforestlabs2024flux1dev}, Nano-Banana-Pro~\citep{gemini3proimage2025}, and GPT-4o (gpt-image-1)~\citep{openai2025_gpt4o_image1}.
Given our focus on recent models, results for older baselines, including SD 1.5, SD 2.1, SD 3.5-Large, and GLIGEN, are presented in the Appendix. 
We also include a prompt-engineering approach for extracting a cognitive map~\citep{cognitive_map} as an additional baseline with GPT-4o.
For comparison with LLM-based editing frameworks, we include SLD and GraPE~\citep{Grape}, both self-correcting pipelines that achieve strong performance using GPT-4o as the LLM interpreter. 
SLD serves as the baseline framework for FoR-SALE, whereas GraPE uses an MLLM to determine editing actions and invokes SOTA image-editing models for modification. We also compare FoR-SALE with a strong image-editing model, Qwen-Image-Edit~\citep{wu2025qwenimagetechnicalreport}.
All experiments were conducted on two A6000 GPUs, totaling approximately 450 GPU-hours. Further implementation details are provided in Appendix~\ref{sec:baseline_details}.

\subsection{FoR-SALE Implementation Details}
We select Qwen3-32B-Reasoning~\citep{Qwen3} for all LLM components in the FoR-SALE pipeline.
For the Visual Perception module, we employ OWLv2~\citep{OWLv2} for object detection, DPT~\citep{DPT} for depth estimation, and OrientAnything~\citep{orient_anything} for orientation detection. 
We use SD 1.5 as the base diffusion model to create objects and as the final step to denoise the composed latent space. 
The same visual processing tools are used in evaluation. To verify that our improvements are not tied to the performance of specific visual processing tools, we additionally report experimental results using alternative ones. The results demonstrate consistent improvements from the FoR-SALE framework, regardless of the choice of vision tools.
This outcome is expected, as the visual tools are limited to basic tasks (e.g., object detection), while the LLM carries out the reasoning.
We report results using the alternative tools in Appendix~\ref{appendix:different_stack_eval}. 
A detailed execution-time analysis of each component is presented in Appendix~\ref{appendix:time_analysis}. Our results show that FoR-SALE introduces only \textbf{a runtime overhead of approximately 45\%} relative to the SLD baseline. This is a reasonable trade-off for enabling spatial reasoning from various perspectives.
More implementation details are provided in Appendix~\ref{sec:appendix_experiment}.

\begin{table*}[t]
    \centering
    \resizebox{0.8\textwidth}{!}{
    \begin{tabular}{l c c c  c c  c  c }
        \toprule
        & \multicolumn{3}{c}{FoR-LMD} & \multicolumn{3}{c}{FoREST} & \\
         \cmidrule(lr){2-4} \cmidrule(lr){5-7}
        Method    
    & \textbf{Rel.} $\uparrow$ 
    & \textbf{Int.} $\uparrow$ 
    & \textbf{Avg.} $\uparrow$
    & \textbf{Rel.} $\uparrow$ 
    & \textbf{Int.} $\uparrow$ 
    & \textbf{Avg.} $\uparrow$ & \textbf{Overall} $\uparrow$  \\
        \midrule
        FLUX.1 & $58.95$ & $25.83$ & $41.00$ & $18.18$ & $15.56$ & $17.00$ & $29.00$ \\
        \ \ \ \ + Qwen-Image-Edit           & $68.12$ & $26.57$ & $45.60$ & $33.45$ & $24.44$ & $29.40$ & $37.50$ \\
        \ \ \ \ + Qwen-Image-Edit (2$\times$) & $68.99$ & $26.57$ & $46.00$ & $\mathbf{34.90}$ & $24.00$ & $30.00$ & $38.00$ \\
        \ \ \ \ + 1-round GraPE & $54.15$ & $18.08$ & $34.60$ & $17.45 $ & $11.56$ & $14.80$ & $24.70$ \\
        \ \ \ \ + 1-round SLD & $63.32$ & $25.09$ & $42.60$ & $24.72 $ & $12.00$ & $19.00$ & $30.80$ \\
        \ \ \ \ \textbf{+ 1-round FoR-SALE} & $65.07$ & $27.67$ & $44.80$ & $25.09$ & $22.22$ & $23.80$ & $34.30$ \\
        \ \ \ \ \textbf{+ 2-round FoR-SALE} & $67.68$ & $\mathbf{28.04}$ & $\mathbf{46.20}$ & $30.18$ & $29.78$ & $30.00$ & $38.10$  \\
        \ \ \ \ \textbf{+ 3-round FoR-SALE} & $\mathbf{69.43}$ & $25.84$ & $45.80$ & ${32.72}$ & $\mathbf{31.11}$ & $\mathbf{32.00}$ & $\mathbf{38.90}$ \\
        \midrule
        Nano-Banana-Pro & $73.36$ & $7.38$ & $37.60$ & $52.00$ & $27.55$ & $41.00$ & $39.30$ \\
        \ \ \ \ + Qwen-Image-Edit & $75.10$ & $9.59$ & $39.60$ & $49.09$ & $26.67$ & $39.00$ & $39.30$ \\
        \ \ \ \ + 1-round GraPE & $73.36$ & $6.64$ & $37.20$ & $48.72$ & $25.33$ & $38.20$ & $37.70$ \\
        \ \ \ \ + 1-round SLD & $81.22$ & $13.28$ & $44.40$ & $53.81$ & $24.44$ & $40.60$ & $42.50$ \\
        \ \ \ \ \textbf{+ 1-round FoR-SALE} & $84.72$ & $14.02$ & $46.40$ & $60.72$ & $28.44$ & $46.20$ & $46.30$ \\
        \ \ \ \ \textbf{+ 2-round FoR-SALE} & $89.52$ & $21.77$ & $52.80$ & $\mathbf{62.18}$ & $29.33$ & $47.40$ & $50.10$ \\
        \ \ \ \ \textbf{+ 3-round FoR-SALE} & $\mathbf{91.70}$ & $\mathbf{26.58}$ & $\mathbf{56.40}$ & $61.45$ & $\mathbf{32.44}$ & $\mathbf{48.40}$ & $\mathbf{52.40}$ \\
        \midrule
        GPT-4o (gpt-image-1) & $\mathbf{94.76}$ & $24.35$ & $56.60$ & $57.81$ & $35.56$ & $47.80$ & $52.20$ \\
        \ \ \ \ + Qwen-Image-Edit & $94.32$ & $22.88$ & $55.60$ & $\mathbf{61.45}$ & $32.00$ & $48.20$ & $51.90$ \\
        \ \ \ \ + 1-round GraPE & $93.89$ & $19.56$ & $53.60$ & $55.64$ & $30.22$ & $44.20$ & $48.90$ \\
        \ \ \ \ + 1-round SLD & $89.08$ & $21.40$ & $52.40$ & $43.27$ & $23.56$ & $34.40$ & $43.40$ \\
        \ \ \ \ \textbf{+ 1-round FoR-SALE} & $93.01$ & $35.42$ & $61.80$ & $54.18$ & $37.33$ & $46.60$ & $54.20$ \\
        \ \ \ \ \textbf{+ 2-round FoR-SALE} & $93.01$ & $34.32$ & $61.20$ & $48.73$ & $39.11$ & $44.40$ & $52.80$ \\
        \ \ \ \ \textbf{+ 3-round FoR-SALE} & $91.26$ & $\mathbf{38.37}$ & $\mathbf{62.60}$ & $53.81$ & $\mathbf{42.22}$ & $\mathbf{48.60}$ & $\mathbf{55.60}$ \\
        \bottomrule
    \end{tabular}
    }
    \caption{Accuracy of generated images across various baselines. Relative (Rel.) denotes camera-perspective spatial expressions, while intrinsic (Int.) denotes spatial expressions from another object’s perspective.} 
    \label{tab:main_result1}
\end{table*}
\subsection{Results and Discussion}\label{sec:experiment_results}

\noindent\textbf{RQ1. Can the SOTA T2I models follow the FoR expressed in the text? }

As shown in Table~\ref{tab:main_result1}, the best-performing model, GPT-4o, achieves only 52.20\% accuracy, \textbf{highlighting the difficulty of T2I generation, even with only two objects in a spatial relation}. While GPT-4o performs well on \textit{relative} FoR in FoR-LMD (94.76\%), its accuracy drops sharply to 24.35\% on \textit{intrinsic} FoR, revealing a substantial performance gap.
A similar trend is observed in a more recent model (Nano-Banana-Pro), which also shows a dramatic drop in intrinsic FoR cases. 
This finding is consistent with prior work~\citep{FoREST, GenSpace}, which highlights the challenges of reasoning about FoR in non-camera perspectives.
Despite their low performance on FoR-LMD, GPT-4o and Nano-Banana-Pro achieve reasonable results on the FoREST benchmark, which explicitly encodes facing direction in the input. 
This is likely due to their 3D spatial training.
In contrast, diffusion models such as FLUX.1 fail to effectively leverage 3D spatial information (e.g., orientation) and struggle under both relative and intrinsic FoR settings in FoREST.
Other diffusion models exhibit behavior similar to that of FLUX.1. Detailed results for these additional baselines are presented in Appendix~\ref{appendix:additional_results}.
In addition to the above baselines, we apply a spatial prompting strategy to extract a cognitive map~\citep{cognitive_map} using GPT-4o.
We find that the cognitive map improves GPT-4o’s object-counting accuracy but provides only limited benefits for FoR spatial reasoning; details are reported in Appendix~\ref{appendix:cognitive_map}.

\noindent\textbf{RQ2. How effective is FoR-SALE in editing images to follow the FoR expressed in text?}

\begin{wrapfigure}{r}{0.4\textwidth}
    \centering
    \vspace{-2mm}
    \includegraphics[width=\linewidth]{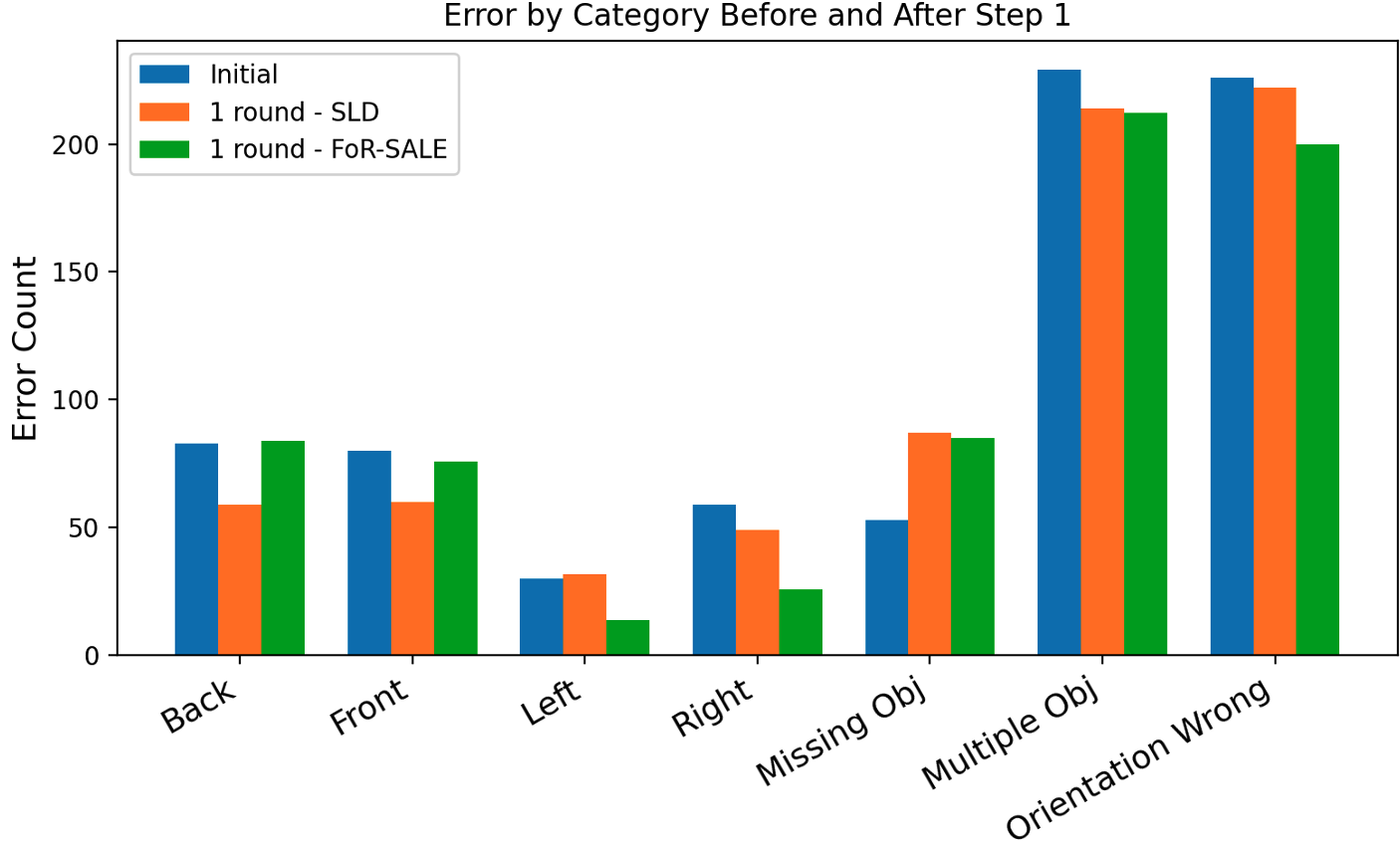}
    \vspace{-6mm}
    \caption{Error analysis of images generated by FLUX.1 (blue), SLD (orange), and FoR-SALE (green).}
    \vspace{-2mm}
    \label{fig:error_FLUX1}
\end{wrapfigure}
To answer this question, we compare FoR-SALE with two self-correcting frameworks, SLD and GraPE, and one SOTA image-editing model, Qwen-Image-Edit, in Table~\ref{tab:main_result1}. 
FoR-SALE generally outperforms both self-correcting frameworks in both relative and intrinsic FoR settings. 
Nevertheless, Qwen-Image-Edit performs better in relative cases when the initial image is generated by FLUX.1.
We attribute this to its strong spatial reasoning ability, which better processes camera-perspective descriptions and significantly improves weaker base models such as FLUX.1. 
However, Qwen-Image-Edit struggles to refine images generated by the stronger baselines, GPT-4o and Nano-Banana-Pro, especially under the considerably more challenging intrinsic-FoR setting.
In contrast, FoR-SALE achieves consistent improvements in intrinsic FoR settings, up to 11.1 percentage points (pp) after one round and 19.2 pp after three rounds, and continues to improve with additional rounds. 
On the other hand, applying additional rounds using Qwen-Image-Edit  yields little to no further improvements. 
Figure~\ref{fig:error_FLUX1} shows the error types in images edited by SLD and FoR-SALE when using initial images from FLUX.1. 
FoR-SALE shows clear improvements in left and right relations, which can often be corrected through 2D spatial adjustments. 
These gains are expected when the Layout Interpreter accurately infers the FoR, highlighting the positive impact of the FoR Interpreter.
FoR-SALE also reduces many orientation errors; however, correcting 3D aspects such as depth and facing direction remains challenging, with high error rates in these categories. Performance on front–back relations shows limited improvement and sometimes worsens relative to SLD, further underscoring the difficulty of 3D editing.
We suspect that SLD’s apparent improvement in front/back errors does not lead to an overall performance increase, as it introduces new errors due to a lack of depth information.
To evaluate this hypothesis, we provide a detailed analysis in 
Appendix~\ref{appendix:error_for_sale}, comparing errors in front and back relations. The analysis reveals that SLD’s front/back errors are reduced due to the generation of extra objects, which are later counted as multiple-object errors. 
We also observe that multiple-object and missing-object errors remain high for both models, highlighting a limitation in current editing frameworks. 
Finally, by sampling failure cases and manually categorizing each error, we find that the majority of mistakes arise from incorrect orientation generation, failures in the final diffusion stage to synthesize the target object, and shortcomings of the Visual Perception Modules in detection. Further details are provided in Appendix~\ref{appendix:error_manual}.

   

%% file: 05_extra_experiment.tex
\begin{table*}[t]
    \centering
    \small
    \setlength{\tabcolsep}{3pt}
    \resizebox{0.85\textwidth}{!}{
        \begin{tabular}{lcccccc}
            \toprule
            \multirow{2}{*}{\textbf{Layout Interpreter}} 
            & \multicolumn{3}{c}{\textbf{LLM-Layout Accuracy}} 
            & \multicolumn{3}{c}{\textbf{Image Accuracy}} \\
            \cmidrule(lr){2-4} \cmidrule(lr){5-7}
            & \textbf{Rel.} $\uparrow$ 
            & \textbf{Int.} $\uparrow$ 
            & \textbf{Avg.} $\uparrow$
            & \textbf{Rel.} $\uparrow$ 
            & \textbf{Int.} $\uparrow$ 
            & \textbf{Avg.} $\uparrow$ \\
            \midrule
            o3 
            & $\mathbf{99.40}$ & $79.03$ & $89.30$ 
            & $69.24$ & $30.64$ & $50.10$ \\
            o4-mini 
            & $99.20$ & $64.52$ & $82.00$ 
            & $\mathbf{74.40}$ & $29.44$ & $52.10$ \\
            \midrule
            Qwen3-32B
            & $98.21$ & $45.97$ & $72.30$ 
            & $73.61$ & $21.77$ & $47.90$ \\
            Qwen3-32B + FoR Interpreter (No Rules) 
            & $95.23$ & $54.03$ & $74.80$ 
            & $69.84$ & $24.80$ & $47.50$ \\
            Qwen3-32B + FoR Interpreter (Partial Rules)  
            & $93.25$ & $81.65$ & $87.50$ 
            & $70.63$ & $\mathbf{39.52}$ & $55.20$ \\
            Qwen3-32B + FoR Interpreter (Generated Rules) 
            & $94.25$ & $82.05$ & $88.20$ 
            & $70.63$ & $38.40$ & $54.60$ \\
            Qwen3-32B + FoR Interpreter (Full Rules)
            & $93.85$ & $84.48$ & $89.20$ 
            & $71.82$ & $36.29$ & $54.20$ \\
            Qwen3-32B + Routing + FoR Interpreter 
            & $98.20$ & $\mathbf{85.68}$ & $\mathbf{92.00}$ 
            & $71.60$ & $39.00$ & $\mathbf{55.70}$ \\
            \bottomrule
        \end{tabular}
    }
    \caption{
    Accuracy of suggested layouts and edited images under different Layout Interpreter settings, using initial images generated by GPT-4o (gpt-image-1). 
    Rel. and Int. denote relative and intrinsic FoR cases, respectively. 
    }
    \label{tab:llm_layout_results}
\end{table*}

\noindent\textbf{RQ3. How accurately do LLMs perform layout editing?}

To address this question, we conduct an ablation study on the LLMs used for the Layout Interpreter, evaluating two SOTA reasoning models: o3 and o4-mini~\citep{openai2025o3o4mini}. 
We also examine three FoR Interpreter settings: (1) No-Rules, where no perspective-conversion rules are provided; (2) Partial-Rules, which include only rules related to facing directions present in the input or detection results; and (3) Full-Rules, which include all manually crafted rules. We report both LLM-generated layout accuracy and final edited-image accuracy using the evaluation protocol in Section~\ref{sec:evaluation_method}.
Table~\ref{tab:llm_layout_results} presents the results. LLM-generated layout accuracy is consistently higher than final-image accuracy, indicating that executing layout-guided edits remains challenging. 
For Qwen3, the FoR Interpreter substantially improves intrinsic-FoR layout accuracy from 45.97\% to 84.48\% with Full-Rules, while Partial-Rules already achieves 81.65\%. This shows that the \textbf{predefined FoR Interpreter rules effectively support intrinsic-FoR reasoning, even when the rule set is incomplete.} With the FoR Interpreter, Qwen3 also outperforms o3 on the more challenging intrinsic-FoR cases.
{However, applying the FoR Interpreter to all prompts slightly reduces Qwen3's relative-FoR accuracy, since relative-FoR prompts do not require perspective conversion. This motivates a routing mechanism that invokes the FoR Interpreter only for intrinsic-FoR cases. Routing improves average layout accuracy from 89.20\% to 92.00\% and increases relative-FoR accuracy from 93.85\% to 98.20\%, while maintaining intrinsic-FoR accuracy comparable to that of the Full-Rules setting. More details about the routing experiment are presented in Appendix~\ref{appendix:routing}.

We further evaluate whether FoR Interpreter rules can be generated automatically by an LLM. The Generated-Rules setting achieves 88.20\% average layout accuracy, close to the 89.20\% achieved with Full-Rules, suggesting that the perspective-conversion rule set can be automatically extended from a small number of examples. We observe a similar trend in the FoREST-G generalization test, where LLM-generated rules remain effective in settings with additional objects and relation types. Experimental details and additional results on FoREST-G are provided in Appendix~\ref{appendix:rule_generation}.
Finally, although o3 produces highly accurate layouts, it yields lower final image accuracy than some alternatives, likely because its layouts require more editing actions; we analyze this further in Appendix~\ref{appendix:error_baseline}.

\noindent\textbf{RQ4. How do the new editing actions help FoR-SALE?} 

To answer this question, we ablate facing-direction and depth modification using FLUX.1-generated images. As shown in Table~\ref{tab:diagnostic_and_ablation}(b), removing facing-direction modification reduces accuracy by 2.8 pp after one round, whereas removing depth modification reduces it by 0.3 pp after one round and approximately 2 pp after two rounds. This shows that facing-direction modification yields a larger immediate gain, whereas depth modification becomes more influential by the second correction round.
We further evaluate a stronger depth-editing backbone by replacing ControlNet with FLUX.1-Fill~\citep{flux1fill}. This backbone improves FoR-SALE's performance, although depth editing remains imperfect, suggesting that more reliable 3D-aware methods could yield further gains. Details of this experiment are provided in Appendix~\ref{appendix:edit_tool_alternative}. Together, these results confirm the importance of both editing operations for spatial alignment.

\begin{table*}[t]
    \centering
    \small
    \renewcommand{\arraystretch}{0.95}
    
    \begin{minipage}[t]{0.45\textwidth}
    \centering
    \resizebox{\linewidth}{!}{
    \begin{tabular}{llccc}
    \toprule
    \textbf{Subset} & \textbf{Model} & \textbf{Rel.}  & \textbf{Int.} & \textbf{Avg.}\\
    \midrule
    \multirow{3}{*}{Complex}
    & FLUX.1 & 8.25 & 8.25 & 8.25 \\
    & Qwen-Image-Edit & \textbf{16.42} & 10.79 & 12.67 \\
    & FoR-SALE & 14.33 & \textbf{12.88} & \textbf{13.37} \\
    \midrule
    \multirow{3}{*}{Rel.-Aug.}
    & FLUX.1 & 14.80 & 10.29 & 12.17 \\
    & Qwen-Image-Edit & \textbf{26.00} & 16.57 & 20.50 \\
    & FoR-SALE & 21.20 & \textbf{21.43} & \textbf{21.33} \\
    \midrule
    \multirow{3}{*}{Lang.-Var.}
    & FLUX.1 & 42.72 & 18.04 & 30.75 \\
    & Qwen-Image-Edit & \textbf{54.37} & 23.20 & 39.25 \\
    & FoR-SALE & 48.54 & \textbf{29.90} & \textbf{39.50} \\
    \bottomrule
    \end{tabular}
}

\vspace{1mm}
\textbf{(a) Diagnostic subsets of FoREST-G.}
\end{minipage}
\hspace{0.03\textwidth}
\begin{minipage}[t]{0.5\textwidth}
\centering
\resizebox{\linewidth}{!}{
    \begin{tabular}{lccc}
    \toprule
    \textbf{Method} & \textbf{Rel.} & \textbf{Int.} & \textbf{Avg.} \\
    \midrule
    FLUX.1 & 38.57 & 20.70 & 29.00 \\
    \quad  + SLD & 42.26 & 19.15 & 30.80 \\
    \midrule
    + FoR-SALE & 43.25 & 25.20 & 34.30 \\
    \quad - Facing Direction Modification & 40.67 & 22.17 & 31.50 \\
    \quad - Depth Modification& 42.65 & 25.20 & 34.00 \\
    \midrule
    + FoR-SALE (2 rounds) & 47.22 & 28.83 & 38.10 \\
    \quad - Depth Modification & 45.83 & 26.21 & 36.10 \\
    \bottomrule
    \end{tabular}
}

\vspace{1mm}
\textbf{(b) Ablation of editing operations.}
\end{minipage}

\vspace{-2mm}
\caption{
Additional diagnostic and ablation results. (a) Image-level accuracy on the diagnostic generalization subsets of FoREST-G. FoR-SALE achieves the highest intrinsic-FoR accuracy across all three subsets. (b) Ablation results using FLUX.1-generated images, demonstrating the contributions of facing-direction modification and depth modification.
}
\label{tab:diagnostic_and_ablation}
\vspace{-2mm}
\end{table*}

\noindent\textbf{RQ5. How does FoR-SALE generalize beyond simple two-object settings?} 

{To address this question, we evaluate FoR-SALE on FoREST-G, which contains three subsets targeting different aspects of generalization: compositional scene structure (Complex), broader relation coverage (Relation-Augmented), and natural language variation (Language-Variation). We compare FoR-SALE with the SOTA image-editing model Qwen-Image-Edit in Table~\ref{tab:diagnostic_and_ablation}(a).
The results show that performance drops substantially on the Complex and Relation-Augmented subsets, reflecting the increased difficulty of handling more complex spatial constraints. Nevertheless, FoR-SALE follows the same trend as on the main benchmarks, achieving the best intrinsic-FoR accuracy across all subsets and the best average accuracy in each subset. This suggests that FoR-SALE remains effective under harder scene structures, expanded relation types, and naturalized prompts, especially for intrinsic-FoR cases, which are the primary challenge in FoR-aware T2I generation.}

%% file: 91_Implementation_Details.tex
\section{FoR-SALE Implementation Details}\label{sec:appendix_experiment}
The random seed is set to 78 in all experiments to ensure reproducibility.

\subsection{LLM Parser}
For the implementation of the LLM Parser, we employ Qwen3-32B with reasoning generation (thinking tokens) disabled to enable faster inference, given the simplicity of the task.
The temperature is set to 0 for reproducible results, and the maximum token limit is 8196.
Listing~\ref{lst:LLM_parser} in Section~\ref{sec:LLM_prompt} provides the complete prompt and examples used for this LLM Parser.

\subsection{FoR Interpreter}
We select Qwen3-32B with reasoning generation (thinking tokens) enabled for the FoR Interpreter, as this component requires reasoning over the provided rules.
To ensure reproducibility, the temperature is set to 0, and the maximum token limit is 8196.
Listing~\ref{lst:FoR_Interpreter} in Section~\ref{sec:LLM_prompt} presents the complete prompt and examples used for the FoR Interpreter.

\subsection{Layout Interpreter}
Similar to the FoR Interpreter, we use Qwen3-32B with reasoning generation (thinking tokens) enabled for this LLM component.
For the ablation study, we also evaluate two additional LLMs via the OpenAI API: o3 (model name: o3-2025-04-16) and o4-mini, both from OpenAI.
To ensure reproducibility, the temperature is set to 0, and the maximum token limit is 8196.
This configuration is applied consistently across all LLMs used in the Layout Interpreter. The prompt for this Layout Interpreter is in Listing~\ref{lst:Layout_Interpreter} in Section~\ref{sec:LLM_prompt}.

\subsection{Visual Perception Module}
For the implementation of the Visual Perception Module, we employ three components, including object detection, depth estimation, and orientation detection, as mentioned in the main paper.
For open-vocabulary object detection, we use OWLv2, with the model ID\textit{ google/owlv2-base-patch16-ensemble}.
For depth estimation, we select DPT, using the model ID \textit{Intel/dpt-large}.
Finally, for orientation detection, we employ OrientAnything, with ViT-Large as the base model. The model weights are loaded from the checkpoint \textit{croplargeEX2/dino\_weight.pt}, as provided in the official GitHub repository.

\subsection{Evaluation Functions}

There are six evaluation functions used to assess the generated image.
The visual details are represented in the following format:
{(\texttt{(attribute) (object name) (\#object ID)}, [$x$, $y$, $w$, $h$], $D_i$, $f_i$)} where ($x$, $y$) indicates the coordinates of the upper-left corner of the bounding box from 0.0 to 1.0, $w$ is its width, $h$ is its height, $D_i$ is depth from 0.0 to 1.0 where 1.0 indicates the nearest position to the camera, and $f_i$ is the facing direction label.
Each comparison involves two objects, denoted as $obj_1$ and $obj_2$.
Before performing the comparison, we compute the center of each object’s bounding box, denoted by ($c_x$, $c_y$), where
$c_x = x + w / 2$ and $c_y = y + h / 2$. The procedure for each comparison is described below.

\begin{itemize}
    \item \textbf{Left}. We determine whether the center of $obj_1$ is to the left of $obj2$ by checking whether $c_{x}$ of $obj_1$ is less than $c_{x}$ of $obj_2$. The condition is defined as follows:
    $$c_x^{obj_1} < c_x^{obj2}$$
    
    \item \textbf{Right}.  We determine whether the center of $obj_1$ is to the right of $ obj_2$ by checking whether $c_{x}$ of $obj_1$ is greater than $c_{x}$ of $obj_2$. The condition is defined as follows: $$c_x^{obj_1} > c_x^{obj2}$$
    
    \item \textbf{Above}. We determine whether the center of $obj_1$ is above $obj_2$ by checking whether $c_{y}$ of $obj_1$ is less than $c_{y}$ of $obj_2$. The condition is defined as follows:
    $$c_y^{obj_1} < c_y^{obj2}$$
    
    \item \textbf{Below}.  We determine whether the center of $obj_1$ is below $ obj_2$ by checking whether $c_{y}$ of $obj_1$ is greater than $c_{y}$ of $ obj_2$. The condition is defined as follows: $$c_y^{obj_1} > c_y^{obj2}$$
    
    \item \textbf{Front}. We determine whether $obj_1$ is in front of $obj_2$ by comparing $D_1$ (depth of $obj1$) with $D_2$ (depth of $obj2$). The condition is defined as follows:
    $$
    D_1 > D_2
    $$
    \item \textbf{Back}. We determine whether $obj_1$ is behind $obj_2$ using the following inequality:
    $$
    D_1 < D_2
    $$
\end{itemize}

\section{FoREST-G Construction}\label{appendix:diag_subset}

{To construct \textbf{FoREST-G}, an extension of FoREST~\citep{FoREST}, we follow the prompt templates from the \textit{Clear} portion of FoREST, where the frame of reference is explicitly specified for each spatial relation. 
We then construct three diagnostic subsets---\textbf{Complex}, \textbf{Relation-Augmented}, and \textbf{Language-Variation}---to evaluate generalization beyond the original two-object setting. Each subset contains 500 spatial expressions, resulting in 1,500 additional examples in total.}

{
\textbf{The Complex subset} introduces chain and star relation topologies to represent relations among multiple objects. Chain structures connect objects sequentially, requiring multi-hop reasoning, while star structures organize multiple relations around a shared reference object. This subset evaluates whether FoR-SALE generalizes beyond isolated pairwise reasoning while retaining the original relation set: \{left, right, front, back\}. We limit each scene to at most four objects.}

{
\textbf{The Relation-Augmented subset} extends the original relation vocabulary beyond \{left, right, front, back\} by introducing diagonal relations (e.g., front-left and back-right) and vertical relations (above and below). To isolate the effect of the expanded relation types, this subset retains the original two-object setting.}

{
\textbf{The Language-Variation subset} follows the original FoREST~\citep{FoREST} construction but uses an LLM to rewrite the generated contexts into more natural and diverse language while preserving the underlying spatial constraints. This subset evaluates generalization beyond template-based spatial descriptions. The prompt used for language rewriting is provided in Listing~\ref{lst:humanized_prompt}.}

\section{Time Efficiency Analysis}\label{appendix:time_analysis}


\begin{table*}[t]
\centering
\small
\renewcommand{\arraystretch}{0.95}
\setlength{\tabcolsep}{6pt}

\begin{minipage}[t]{0.5\textwidth}
\centering
\begin{tabular}{lc}
\toprule
\textbf{FoR-SALE Component} & \textbf{Time (s)} \\
\midrule
LLM Parser & 21.64 \\
Object Detection & 0.94 \\
Depth Estimation & 1.23 \\
Orientation Detection & 1.60 \\
FoR Interpreter & 31.84 \\
Layout Interpreter & 45.52 \\
Latent Operations & 17.91 \\
Final Image Generation & 7.99 \\
\bottomrule
\end{tabular}

\vspace{1mm}
\textbf{(a) FoR-SALE runtime breakdown.}
\end{minipage}
\hfill
\begin{minipage}[t]{0.4\textwidth}
\centering
\begin{tabular}{lc}
\toprule
\textbf{Method} & \textbf{Time (s)} \\
\midrule
FoR-SALE & 128.67 \\
SLD & 88.52 \\
Qwen-Image-Edit & 145.81 \\
\bottomrule
\end{tabular}

\vspace{1mm}
\textbf{(b) End-to-end runtime.}
\end{minipage}

\vspace{-1mm}
\caption{
Runtime analysis. 
(a) Execution time of each FoR-SALE component.
(b) Average end-to-end runtime of editing methods under the same setting.
}
\label{tab:time_comsumption}
\vspace{-2mm}
\end{table*}

We report the execution time of each component in Table~\ref{tab:time_comsumption}, averaged over one round of FoR-SALE applied to an initial baseline SLD pipeline. Note that the baseline SLD in this experiment uses Qwen3-32B as the LLM-Interpreter to ensure a fair comparison with FoR-SALE.
The overhead is primarily attributable to three LLM-based components. The LLM Parser, which extracts objects from the input, is invoked only in the first round and therefore introduces no additional cost in subsequent editing rounds. 
The Layout Interpreter is the most time-consuming component, as it requires the LLM to reason about and propose a spatial layout. 
This cost is comparable to that of the original SLD framework, which similarly relies on both a parser and the Layout Interpreter.
Overall, FoR-SALE introduces a runtime overhead of approximately 45\% relative to SLD, with the FoR Interpreter accounting for most of the additional cost. 
This trade-off is justified by the ability to handle spatial relations from non-camera perspectives, a challenging yet largely overlooked aspect in prior work, and yields a notable accuracy improvement even after a single round of FoR-SALE under our evaluation benchmarks.

\section{Baseline Model Parameters}\label{sec:baseline_details}

\subsection{Stable Diffusion (SD)}
For baselines using SD1.5 and SD2.1, we set the number of inference steps to 50, while keeping all other parameters at their default values.
The model ID for SD1.5 is \textit{sd-legacy/stable-diffusion-v1-5}, and for SD2.1, it is \textit{stabilityai/stable-diffusion-2-1}.
The baseline using SD3.5-Large employs the model ID \textit{stabilityai/stable-diffusion-3.5-large}, with the number of inference steps set to 30; all other parameters remain unchanged.

\subsection{GLIGEN}
We use Qwen3 to generate the initial layout for the GLIGEN baseline. The prompt used for layout generation is shown in Listing~\ref{lst:LLM_GLIGEN}.
For the GLIGEN model, we use the model ID \textit{masterful/gligen-1-4-generation-text-box}. 
We also provide facing direction information when generating images with GLIGEN by augmenting the object names with the corresponding facing directions extracted from the layout generated by Qwen3.
The number of inference steps is set to 50, while all other parameters remain unchanged.

\subsection{FLUX.1}
For generating images with the FLUX.1 baseline, we employ the pipeline with model ID \textit{black-forest-labs/FLUX.1-dev}. The guidance scale is set to 3.5, following the recommended value. The image resolution is 1024×1024, and the number of inference steps is set to 50. Other parameters are set to the default values.

\subsection{GPT-4o (gpt-image-1)}
We utilize the OpenAI API to generate images for the GPT-4o baseline, employing the model ID \texttt{gpt-image-1}.
The background setting is set to auto, and the image resolution is configured to 1024×1024. All other parameters are left at their default values.
The observed API cost per generated image in our experiments was approximately $\$0.01 - \$0.02$.

\subsection{Nano-Banana-Pro}
We use the Gemini API to generate images with Nano-Banana-Pro, using the model ID \texttt{gemini-3-pro-image-preview}. We set only the image aspect ratio to \textbf{1:1}, while keeping all other parameters at their default values.

\subsection{Qwen-Image-Edit}
For image editing with Qwen-Image-Edit, we employ the diffusion pipeline with the model ID \texttt{Qwen/Qwen-Image-Edit}. We use the following prompt to guide the editing process to match the spatial expression: “\textit{Edit the given image to match the following description: }$T$,” where $T$ denotes the input spatial expression. The number of inference steps is set to 30, following the model’s recommendation. All other parameters are kept at their default values.


%% file: 92_Additional_results.tex
\section{Additional Results on Text-to-Image (T2I) baselines}\label{appendix:error_baseline}

\begin{table*}[t]
    \centering
    \small
    \resizebox{\textwidth}{!}{
    \begin{tabular}{l c c c  c c  c  c }
        \toprule
        & \multicolumn{3}{c}{FoR-LMD} & \multicolumn{3}{c}{FoREST} & \\
        \cmidrule(lr){2-4}  \cmidrule(lr){5 - 7} 
        Method & Relative & Intrinsic & Average & Relative & Intrinsic & Average & Overall Avg.  \\
        \midrule
        SD 1.5 & $12.66$ & $11.80$ & $12.20$ & $7.63$ & $4.00$ & $6.00$ & $9.10$ \\
        SD 2.1  & $13.97$ & $10.33$ & $12.00$ & $5.09$ & $7.11$ & $6.00$ &  $9.00$ \\
        Qwen3 + GLIGEN & $58.52$ & $21.40$ & $38.40$ & $2.54$ & $1.33$ & $2.00$ & $20.20$ \\
        \midrule
               SD 3.5-Large & $63.75$ & $24.72$ & $42.60$ & $18.11$ & $11.11$ & $15.00$ & $28.80$ \\
        \ \ \ \ + 1-round GraPE & $55.46$ & $16.97$ & $34.60$ & $14.91$ & $7.56$ & $11.60$ & $23.10$  \\
        \ \ \ \ + 1-round SLD & $61.57$ & $19.56$ & $38.80$ & $22.55$ & $11.55$ & $17.60$ & $28.20$  \\
        \ \ \ \ \textbf{+ 1-round FoR-SALE} & $61.14$ & $26.56$ & $42.40$ & $24.00$ & $16.00$ & $20.40$ & $31.40$ \\
        \ \ \ \ \textbf{+ 2-round FoR-SALE} & $67.25$ & $26.94$ & $45.40$ & $28.00$ & $22.22$ & $25.40$ & $35.40$ \\
        \ \ \ \ \textbf{+ 3-round FoR-SALE} & $\mathbf{70.31}$ & $\mathbf{29.52}$ & $\mathbf{48.20}$ & $\mathbf{28.00}$ & $\mathbf{22.22}$ & $\mathbf{25.40}$ & $\mathbf{36.80}$ \\
        \midrule
        GPT-4o & $\mathbf{94.76}$ & $24.35$ & $56.60$ & $\mathbf{57.81}$ & $35.56$ & $47.80$ & $52.20$ \\
        \ \ \ \ + Cognitive Map & $98.23$ & $30.36$ & $62.00$ & $56.00$ & $37.33$ & $47.60$ & $54.80$ \\
        \ \ \ \ \textbf{+ 1-round FoR-SALE} & $93.01$ & $35.42$ & $61.80$ & $54.18$ & $37.33$ & $46.60$ & $54.20$ \\
        \bottomrule  
    \end{tabular}
    }
    \caption{Accuracy of images generated by additional baselines and FoR-SALE. Relative denotes camera-based spatial expression; Intrinsic uses another object’s perspective.}
    \label{tab:main_result2}
\end{table*}

\begin{table*}[t]
    \centering
    \small
    \begin{tabular}{l c c c} \\ \toprule  
    Error Type & Missing Object & Over-Generated Objects & Spatial Relations \\ \midrule
    Cognitive Map & $7$ & $38$ & $173$\\  
    FoR-SALE & $10$ & $75$ & $177$ \\ 
    \bottomrule
    \end{tabular}
    \caption{Number of errors in each category obtained using the cognitive map and FoR-SALE using GPT-4o as the base generator.}
    \label{tab:cognitive_map_error}
\end{table*}

\subsection{Additional Results for Early T2I Models}\label{appendix:additional_results}

We provide additional results for early T2I models, including SD1.5, SD2.1, and GLIGEN, using layouts generated by Qwen3, as shown in Table~\ref{tab:main_result2}.
All models perform significantly worse than the SOTA baselines discussed in the main results—particularly SD1.5 and SD2.1, which achieve less than 10\% accuracy.
While GLIGEN shows more acceptable performance on the FoR-LMD benchmark, it performs poorly when orientation requirements are introduced, as in the context of the FoREST benchmark.
GLIGEN’s accuracy drops to just 2\%, indicating a lack of understanding of object-level attributes—especially facing direction—even when this information is explicitly provided during generation. For SD3.5-Large, the model exhibits behavior similar to FLUX.1 in Table~\ref{tab:main_result1}, showing strong performance on relative FoR; however, it performs poorly on intrinsic FoR and when processing 3D spatial information.

\subsection{Image Generation Errors across Different Baselines}

\begin{figure}
    \centering
    \includegraphics[width=0.6\linewidth]{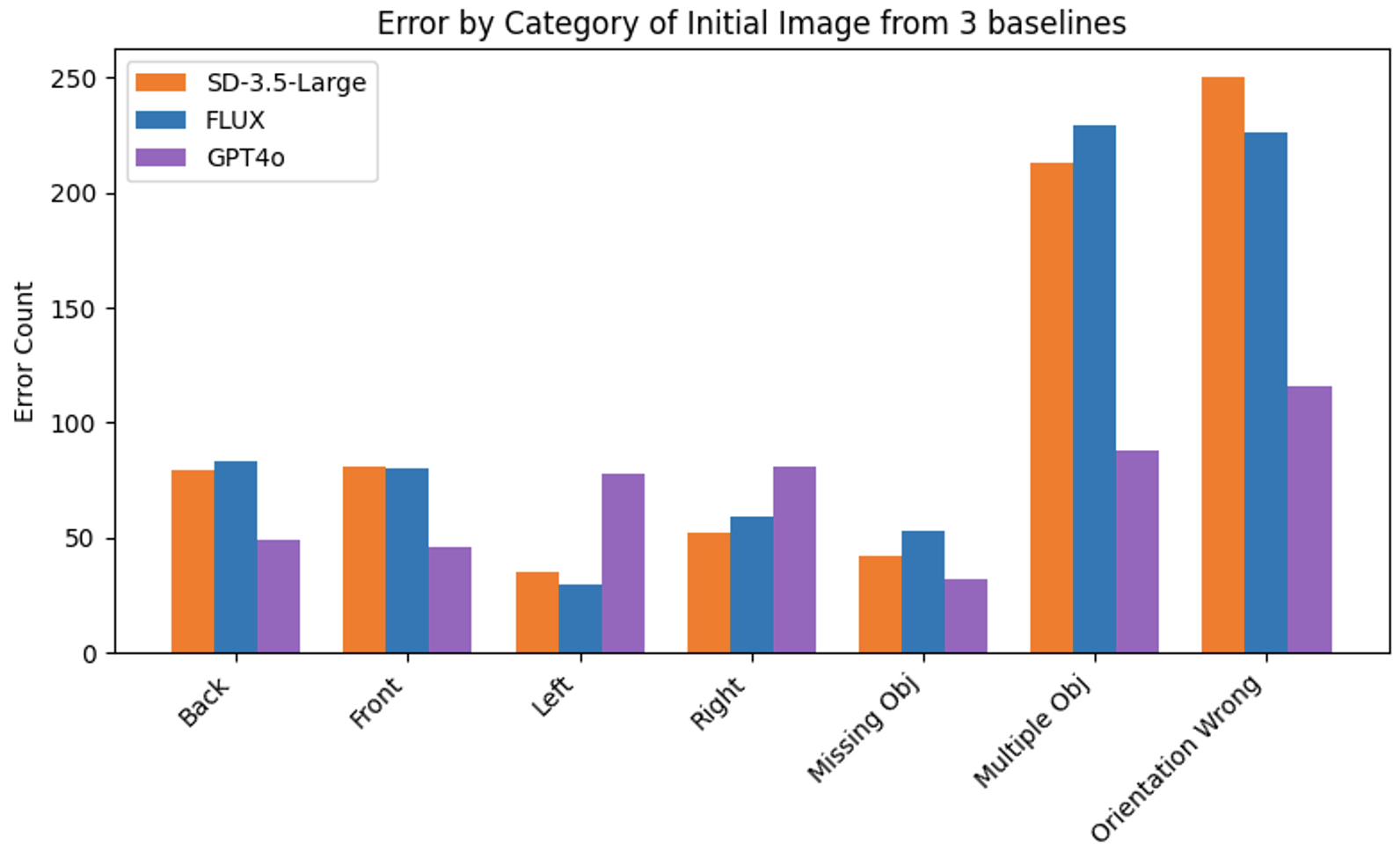}
    \caption{Error analysis of images generated by SD 3.5-Large, FLUX.1, and GPT-4o.}
    \label{fig:error_initial}
\end{figure}

Figure~\ref{fig:error_initial} illustrates the error distribution for images generated by SD3.5-Large, FLUX.1, and GPT-4o.
We observe notable differences among these models. Note that, while SD3.5-Large and FLUX.1 are diffusion-based T2I models, GPT-4o is a unified generative model trained on multimodal input-output tasks.
GPT-4o exhibits significantly fewer missing or additional key objects, indicating stronger object grounding and a more accurate object count.
It also shows lower error rates in front/back relations and orientation, suggesting improved performance in handling 3D spatial configurations, including depth and facing direction.
However, GPT-4o performs worse on left/right relations compared to the diffusion-based models.
We anticipate that this may be attributed to challenges in perspective conversion, as evidenced by GPT-4o’s high performance on relative FoRs in the FoR-LMD benchmark (94.76\%), which requires only camera-centric understanding, contrasted with its significantly lower accuracy on intrinsic FoRs, as reported in the main results.
These findings suggest a trade-off in GPT-4o’s spatial performance—namely, strong handling of camera-centric spatial expressions, but limited generalization to non-camera perspectives in text-to-image tasks.

\begin{table*}[t]
    \centering
    \small
    \setlength{\tabcolsep}{2pt}
    \renewcommand{\arraystretch}{0.9}
    \begin{tabular}{lcccccccc}
    \toprule
    \textbf{Layout Interpreter} 
    & \textbf{Add} 
    & \textbf{Remove} 
    & \textbf{Attr.} 
    & \textbf{Reposition} 
    & \textbf{Facing} 
    & \textbf{Depth} 
    & \textbf{Depth + Rep.} 
    & \textbf{\# Actions} \\
    \midrule
    o3  
    & 3.60 & 10.63 & 0.00 & 49.82 & 15.95 & 10.45 & 9.55 & 1110 \\
    o4-mini 
    & 4.98 & 11.92 & 0.00 & 39.85 & 20.64 & 18.98 & 3.75 & 906 \\
    \midrule
    Qwen3 
    & 3.66 & 10.47 & 0.00 & 36.65 & 16.86 & 25.65 & 6.70 & 955 \\
    \quad + FoR-I (No Rules) 
    & 3.81 & 9.18 & 0.00 & 42.47 & 13.40 & 26.91 & 4.23 & 970 \\
    \quad + FoR-I (Partial Rules) 
    & 3.33 & 8.22 & 0.00 & 41.78 & 10.76 & 31.12 & 4.79 & 1022 \\
    \quad + FoR-I (Full Rules) 
    & 3.40 & 6.90 & 0.00 & 43.25 & 11.95 & 29.74 & 4.86 & 1029 \\
    \bottomrule
    \end{tabular}
    \caption{
    Distribution of editing actions required for FoR-LMD and FoREST using GPT-4o-generated initial images under different Layout Interpreter settings. FoR-I denotes FoR Interpreter. Attr., Facing, Depth, and Rep. denote Attribute Modification, Facing Direction Modification, Depth Modification, and Reposition, respectively.
    }
    \label{tab:action_count}
\end{table*}

\subsection{Cognitive Map}\label{appendix:cognitive_map}


{
We provide additional results using cognitive-map prompting, a recent spatial-reasoning approach~\citep{cognitive_map}. We use prompt engineering to extract a cognitive map before generating the image with GPT-4o, rather than applying one round of FoR-SALE.
We provide the results in Table~\ref{tab:main_result2}.
We observe that the model achieves a significant gain on FoR-LMD, especially on the relative FoR. This is expected, as the cognitive map should improve 2D spatial relations. We further compared the errors in FoR-LMD based on the output of our automatic evaluator to investigate the source of this improvement in Table~\ref{tab:cognitive_map_error}. We found that the cognitive map helps the model predict a more accurate count of objects, which contributes to improved FoR-LMD in both relative and intrinsic FoR. 
However, cognitive-map prompting remains weaker on intrinsic FoR-LMD than FoR-SALE (30.36\% versus 35.42\%), indicating that improved object counting does not resolve intrinsic-FoR reasoning. This is where our proposed framework provides the most value.}

\subsection{FoR Interpreter Routing}\label{appendix:routing}

As discussed in Section~\ref{sec:experiment_results}, the FoR Interpreter can slightly hurt performance on spatial expressions with relative FoR. To mitigate this issue, we introduce a routing mechanism that determines whether the FoR Interpreter should be invoked based on FoR classification.

In this experiment, we use an LLM to classify whether the prompt expresses an intrinsic or relative spatial relation. If the prompt contains an intrinsic spatial relation, the pipeline invokes the FoR Interpreter before the Layout Interpreter. Otherwise, for relative/camera-centric relations, the pipeline bypasses the FoR Interpreter and directly invokes the Layout Interpreter. This prevents unnecessary FoR conversion for prompts that are already expressed in a relative/camera-centric frame. The prompt used for this classification is provided in Listing~\ref{lst:routing_ex}.

Based on the results in Table~\ref{tab:for_extensibility}, the routing mechanism improves average accuracy from 89.20 to 92.00 over Qwen3 + FoR Interpreter. It also recovers the relative-FoR performance drop caused by unnecessary FoR conversion, increasing relative accuracy from 93.85 to 98.20. Intrinsic-FoR performance remains comparable to the non-routed FoR Interpreter setting as expected, since intrinsic prompts are still handled by the FoR Interpreter.

\begin{table}[t]

\centering

\small

\setlength{\tabcolsep}{1pt}

\begin{tabular}{llccc}

\toprule

\textbf{Evaluation} & \textbf{Model} & \textbf{Relative} $\uparrow$ & \textbf{Intrinsic} $\uparrow$ & \textbf{Avg.} $\uparrow$ \\

\midrule

\multirow{5}{*}{FoR-LMD \& FoREST}

& o3 & \textbf{99.40} & 79.03 & 89.30 \\

& Qwen3-32B & 98.21 & 45.97 & 72.30 \\

& Qwen3-32B + FoR Interpreter (32 rules) & 93.85 & 84.48 & 89.20 \\

& Qwen3-32B + FoR Interpreter (Generated rules) & 94.25 & 82.05 & 88.20 \\

& Qwen3-32B + Routing + FoR Interpreter & 98.20 & \textbf{85.68} & \textbf{92.00} \\

\midrule

\multirow{3}{*}{Complex (FoREST-G)}

& o3 & \textbf{97.59} & 74.43 & \textbf{85.67} \\

& Qwen3-32B & 88.65 & 56.63 & 72.17 \\

& Qwen3-32B + FoR Interpreter (32 rules) & 82.47 & \textbf{76.70} & 79.50 \\

\midrule

\multirow{4}{*}{Rel.-Aug. (FoREST-G)}

& o3 & \textbf{97.28} & 75.16 & \textbf{86.00} \\

& Qwen3-32B & 87.76 & 50.33 & 68.67 \\

& Qwen3-32B + FoR Interpreter (32 rules) & 73.47 & 75.82 & 74.67 \\

& Qwen3-32B + FoR Interpreter (Generated rules) & 78.23 & \textbf{79.08} & 78.67 \\

\midrule

\multirow{3}{*}{Lang.-Var.(FoREST-G)}

& o3 & \textbf{100.00} & 86.60 & \textbf{93.50} \\

& Qwen3-32B & 97.09 & 52.58 & 75.50 \\

& Qwen3-32B + FoR Interpreter (32 rules) & 92.23 & \textbf{92.78} & 92.50 \\

\bottomrule

\end{tabular}

\caption{
Layout accuracy under different FoR Interpreter settings on the original and diagnostic evaluation subsets.
}

\label{tab:for_extensibility}

\end{table}

\subsection{FoR Interpreter Additional Experiment}\label{appendix:rule_generation}

\paragraph{LLM as Rule Generator.} To examine the scalability of the handcrafted FoR rules presented in Appendix~\ref{appendix:rules}, we conduct additional experiments in which an LLM generates perspective-conversion rules from few-shot examples. We consider two settings.

First, we provide only four example rules and prompt an LLM, o3~\citep{openai2025o3o4mini}, to generate the remaining rules for the benchmark relations and facing directions used in FoR-LMD and FoREST. This setting tests whether the LLM can recover the original rule space from partial examples. As shown in Tables~\ref{tab:llm_layout_results} and~\ref{tab:for_extensibility}, the generated rules provide sufficient information for the FoR Interpreter, achieving performance close to the manually crafted rules in the original setting (FoR-LMD and FoREST). This suggests that the rules can be generated and used effectively.

Second, we provide all 32 original rules and ask the LLM to generate rules for the new relations in the Relation-Augmented subset of FoREST-G. This setting tests whether the FoR Interpreter can extend beyond the original left, right, front, and back relations. We then augment the FoR Interpreter with the generated rules and evaluate its performance on this subset. As shown in Table~\ref{tab:for_extensibility}, the 32-rule FoR Interpreter improves intrinsic accuracy from 50.33 to 75.82 over vanilla Qwen3, showing that explicit FoR conversion remains useful beyond the original benchmark setting. Adding LLM-generated rules further improves intrinsic accuracy to 79.08 and average accuracy to 78.67, yielding an average gain of 4.0 pp over the 32-rule variant. This suggests that the rule set can be extended beyond the original relations, and that rules generated from the handcrafted examples remain useful for more complex relation types.

\paragraph{FoREST-G.} As shown in Table~\ref{tab:for_extensibility}, we evaluate the FoR Interpreter on FoREST-G using GPT-4o-generated initial images. Across all FoREST-G subsets, which test generalization to complex scenes, augmented relation types, and language variation, incorporating the FoR Interpreter consistently improves Qwen3, especially on intrinsic-FoR cases. In several settings, Qwen3 with the FoR Interpreter approaches or matches the performance of the stronger o3 model. These results highlight the importance of explicit FoR conversion in helping LLMs interpret spatial relations from different frames of reference.

\section{Visual Perception Modules}\label{appendix:different_stack_eval}

In this section, we present additional experiments with alternative visual perception stacks to ensure that the observed improvements are not tied to the specific perception modules used in FoR-SALE or in the evaluation protocol. Specifically, we replace the object detector with Grounding-DINO~\citep{grounding_dino} and the depth estimator with MiDaS 3.0~\citep{midas3-1}. We then re-evaluate the same images generated by FLUX.1 and one-round FoR-SALE, using the initial FLUX.1 images as input.
The results are reported in Table~\ref{tab:different_stack_eval}. We observe that both the original evaluation stack in Table~\ref{tab:main_result1} and the alternative stack in Table~\ref{tab:different_stack_eval} show similar improvements of 5.3 and 6.7 pp, respectively, over the initial FLUX.1 images, a difference of only 1.4 pp between the two evaluation settings. These findings indicate that FoR-SALE consistently improves performance even when different visual perception modules are used.

\begin{table}[t]
    \centering
    \small
    \setlength{\tabcolsep}{4pt}
    \begin{tabular}{lcccccc}
    \toprule
    \multirow{2}{*}{\textbf{Method}} 
    & \multicolumn{3}{c}{\textbf{FoR-LMD}} 
    & \multicolumn{3}{c}{\textbf{FoREST}} \\
    \cmidrule(lr){2-4} \cmidrule(lr){5-7}
    & \textbf{Rel.} $\uparrow$ 
    & \textbf{Int.} $\uparrow$ 
    & \textbf{Avg.} $\uparrow$
    & \textbf{Rel.} $\uparrow$ 
    & \textbf{Int.} $\uparrow$ 
    & \textbf{Avg.} $\uparrow$ \\
    \midrule
    FLUX.1 
    & 62.44 & 33.21 & 46.60 
    & 19.27 & 12.00 & 16.00 \\
    
    Qwen-Image-Edit 
    & \textbf{72.05} & 35.42 & 52.20 
    & 32.00 & 20.89 & 27.00 \\
    
    Qwen-Image-Edit (2$\times$) 
    & 71.61 & 35.05 & 51.80 
    & \textbf{32.36} & 21.78 & 27.60 \\
    
    \midrule
    + FoR-SALE (1 round) 
    & 66.81 & 41.69 & 53.20 
    & 23.63 & 21.78 & 22.80 \\
    
    + FoR-SALE (2 rounds) 
    & 69.43 & \textbf{43.91} & 55.60 
    & 28.36 & 24.00 & 26.40 \\
    
    + FoR-SALE (3 rounds) 
    & 71.61 & \textbf{43.91} & \textbf{56.20} 
    & 31.27 & \textbf{28.89} & \textbf{30.20} \\
    \bottomrule
    \end{tabular}
    \caption{
    Accuracy of generated images with an alternative perception stack. We replace OWLv2 and DPT with Grounding-DINO and MiDaS, respectively. The improvement trend remains consistent, especially for intrinsic-FoR cases.
    }
\label{tab:different_stack_eval}
\end{table}

\section{Analysis of the FoR-SALE Framework}\label{appendix:error_for_sale}

\begin{figure}
    \centering
    \includegraphics[width=0.5\linewidth]{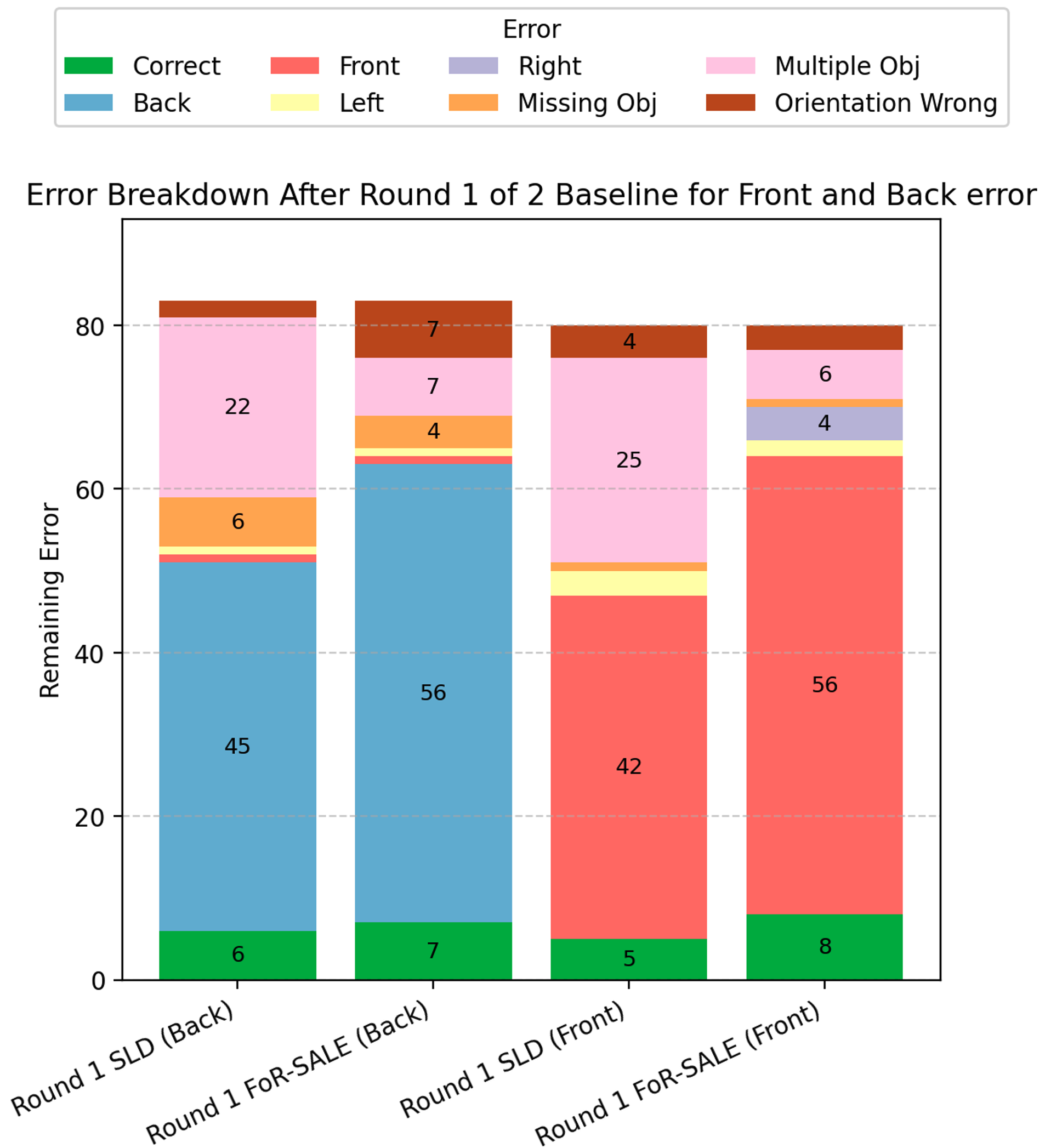}
    \caption{Error breakdown after one round of editing initial images from FLUX.1 using SLD and FoR-SALE on front and back relation errors.}
    \label{fig:error_front_back_breakdown}
\end{figure}

\subsection{Additional Quantitative Error Analysis of FoR-SALE}
In this section, we further analyze the errors observed after applying one round of FoR-SALE to the initial images generated by FLUX.1.
We compare SLD and FoR-SALE in editing images containing front/back spatial relation errors in Figure~\ref{fig:error_front_back_breakdown}.
We observe that while SLD attempts to correct the front/back relation, it often introduces multiple instances of the target objects instead of editing the original ones.
This behavior results in a lower front/back error after one round of editing, but it comes at the cost of generating additional object-related errors.
We attribute this limitation to SLD’s lack of depth awareness, which leads to incorrect editing operations.
In contrast, FoR-SALE, which incorporates depth information, achieves slightly better correction on front/back errors without introducing new object duplication or misalignment.
Importantly, FoR-SALE avoids introducing new error types, making it more robust for subsequent editing rounds.

\subsection{Detailed Analysis of the Effects of Different Layout Interpreters and Editing Actions}
We report the distribution of editing actions required for images generated by GPT-4o when using different Layout Interpreters in Table~\ref{tab:action_count}.
We observe that o3 requires significantly more editing actions compared to other models, with repositioning accounting for 59.37\% of all actions (repositioning and depth modification with repositioning).
This suggests that o3 often generates layouts where the object is repositioned, likely indicating that it is proposing an entirely new scene layout rather than minimally adjusting the original.
This behavior may explain the performance drop observed when using o3-generated layouts, as reported in the main results.
It also highlights a limitation of the FoR-SALE framework: the difficulty in handling cases that require multiple or complex repositioning actions.
These findings suggest that future work may explore improved strategies for accurately moving objects—or even fully regenerating images—when layout revisions are extensive.

\subsection{Alternative Depth Editing Tools}\label{appendix:edit_tool_alternative}

To investigate whether the depth editing error is specific to ControlNet rather than the FoR reasoning component, we conduct an additional experiment replacing ControlNet with FLUX.1-Fill-Dev as a stronger recent editing backbone on the FoREST benchmark.

According to Table~\ref{tab:alter_edit_tool}, the results show that FLUX.1-Fill-Dev improves final image-level accuracy for both relative and intrinsic FoR cases. Relative accuracy increases from 25.09\% to 31.27\%, and intrinsic accuracy increases from 22.22\% to 28.44\%. The error analysis also shows reductions in key execution-level failure modes: missing-object errors decrease from 8.80\% to 8.20\%, and orientation errors decrease from 40.00\% to 33.60\%. However, not all failure modes improve. 3D incorrectness slightly increases from 7.00\% to 7.40\%, and extra-object errors increase from 17.60\% to 18.00\%. This suggests that FLUX.1-Fill-Dev better handles orientation editing, but can introduce other generation errors, especially object-count inconsistency.

Nevertheless, the absolute accuracy remains limited, reflecting the difficulty of the evaluated dataset rather than a failure of the FoR reasoning module. 
The task requires editors to jointly modify orientation, preserve plausible depth, and satisfy the target FoR relation.
The low FoREST performance of Qwen-Image-Edit, despite being a strong recent editor, further suggests that these failures stem from general limitations of current diffusion/editing backbones on orientation-sensitive spatial editing rather than from FoR-SALE itself.

Overall, this experiment shows that FoR-SALE is modular and can benefit from stronger editing backbones, as reflected by improved relative/intrinsic accuracy and reduced orientation errors with FLUX.1-Fill-Dev. At the same time, fully reliable image-level FoR correction remains constrained by the capabilities of current editing models.
\begin{table}[t]
\centering
\small
\setlength{\tabcolsep}{5pt}

\begin{tabular}{lcccccc}

\toprule
\textbf{Model} & \textbf{Rel.} $\uparrow$  & \textbf{Int.} $\uparrow$  & \textbf{3D Inc.} $\downarrow$  & \textbf{Missing} $\downarrow$ & \textbf{Extra} $\downarrow$ & \textbf{Orient. Err.} $\downarrow$\\

\midrule

FLUX.1 & 18.18 & 15.56 & -- & -- & -- & -- \\ 

FoR-SALE & 25.09 & 22.22 & 7.00\% & 8.80\% & 17.60\% & 40.00\% \\

FoR-SALE + FLUX.1-Fill-Dev & \textbf{31.27} & \textbf{28.44} & 7.40\% & \textbf{8.20\%} & 18.00\% & \textbf{33.60\%} \\

\bottomrule

\end{tabular}

\caption{
Results with FLUX.1-Fill-Dev as a stronger fill-based editing backbone compared with the original baseline. 3D Inc. denotes incorrect 3D relations, Missing denotes missing-object errors, Extra denotes extra-object errors, and Orient. Err. denotes orientation errors.
}
\label{tab:alter_edit_tool}

\end{table}

\section{Failure Case Observation}\label{appendix:error_manual}
\begin{figure*}[t]
    \centering
    \includegraphics[width=\linewidth]{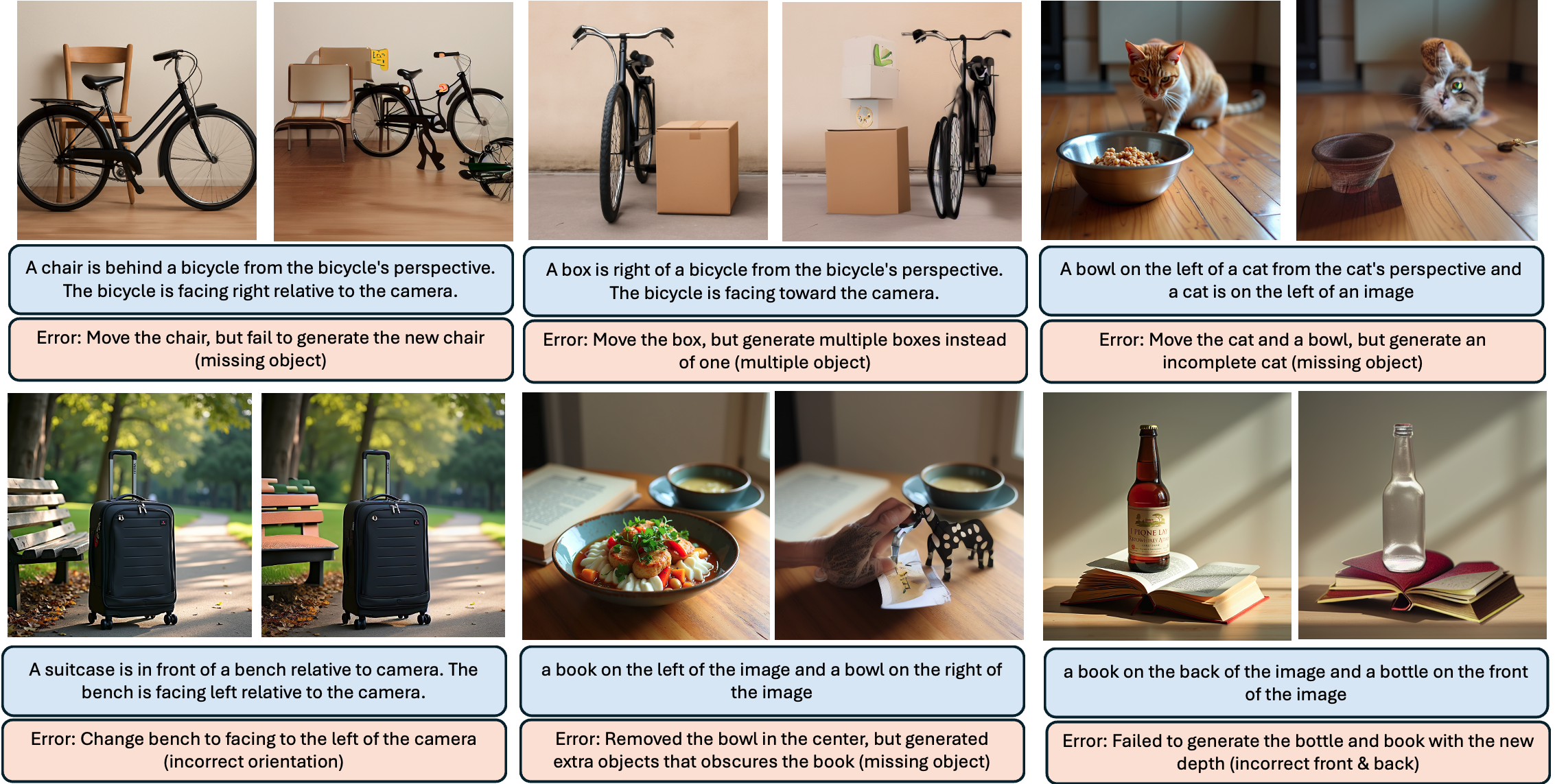}
    \caption{Examples of editing errors using FoR-SALE. The blue box indicates the input spatial expression, while the orange box explains the editing action and the underlying reason for the error.}
    \label{fig:overall_fail}
\end{figure*}

We present examples of FoR-SALE editing failures in Figure~\ref{fig:overall_fail}. The most common errors include multiple
instances of key objects, incorrect orientation,
and missing objects, as also reflected in the main paper’s quantitative results.
We attribute these failures primarily to challenges in object removal and regeneration, which can lead to either the unintended deletion of key objects or the generation of extraneous ones—ultimately making the intended objects undetectable in the final image.
Additionally, we believe that modifying orientation and depth remains difficult for current diffusion models, which limits the effectiveness of FoR-SALE in correcting these types of spatial errors.

\subsection{Percentage of FoR-SALE failure cases}

\noindent\textbf{Observation Setting.} To identify the sources of FoR-SALE generation errors, we sample 60 images (10\% of failure cases in round 1) from the incorrect cases produced by applying one round of FoR-SALE to initial images generated by FLUX.1.
We categorize the errors into six types:

\noindent\textbf{1. Removing object failure (E1).} FoR-SALE fails to remove an object from the image completely.  
    
\noindent\textbf{2. Incorrect orientation generation (E2).} The diffusion model in FoR-SALE generates objects with incorrect orientations.  

\noindent\textbf{3. Incorrect depth generation (E3).} The diffusion model with ControlNet in FoR-SALE generates images with incorrect depth. 

\noindent\textbf{4. Incorrect object generation (E4).} The final Stable Diffusion model in FoR-SALE fails to synthesize the correct object image from the edited latent space.  

\noindent\textbf{5. Object detection failure (E5).} The visual perception modules fail to detect correct object properties or spatial coordinates. 

\noindent\textbf{6. LLM failure (E6).} The LLM-controlled editing suggests an incorrect image layout, leading to erroneous edits.

\begin{figure}[t]
    \centering
    \includegraphics[width=0.5\linewidth]{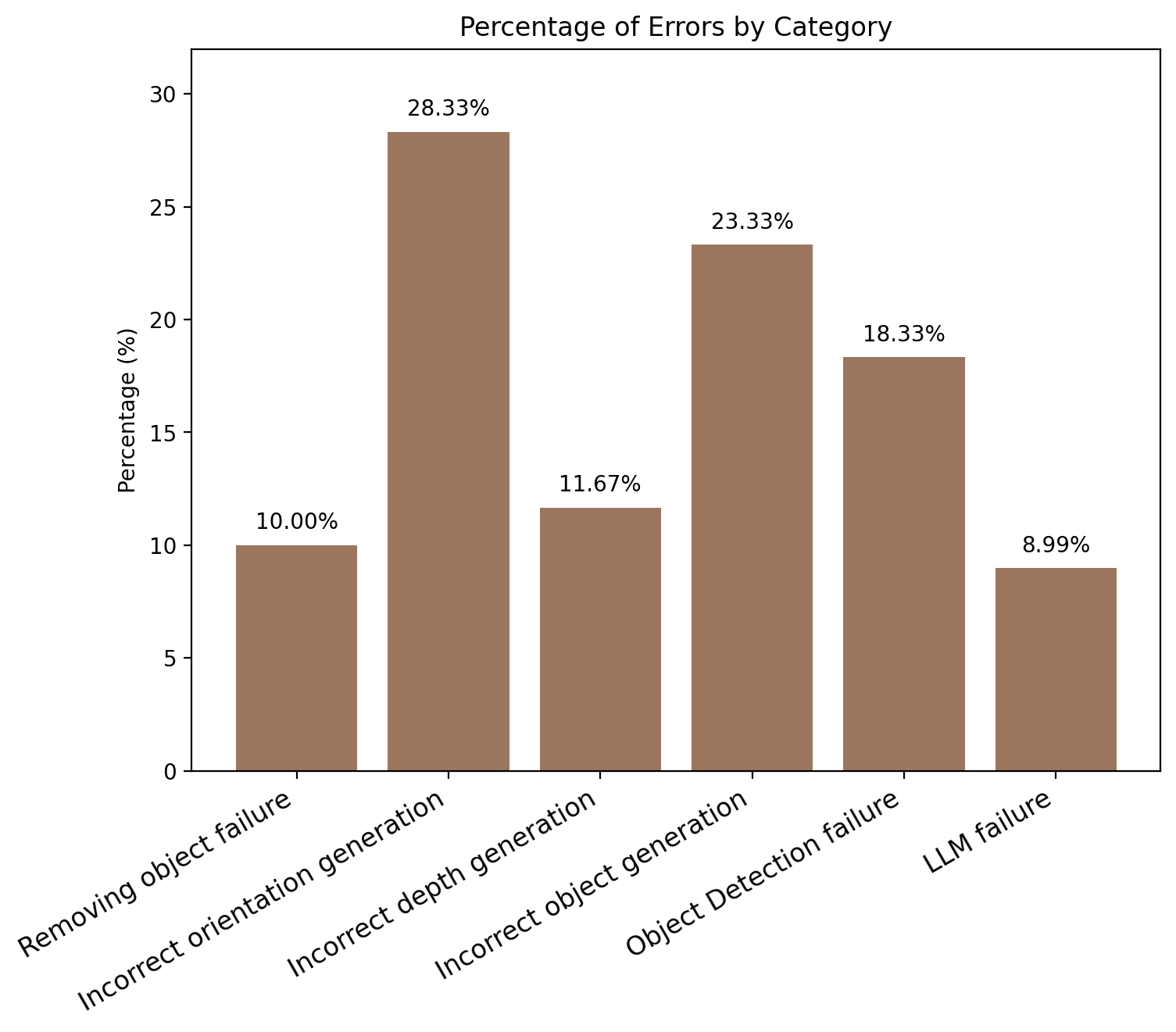}
    \caption{Percentage of each error category among 60 sampled failure cases after one round of FoR-SALE applied to FLUX.1-generated images.}
    \label{fig:error_percentage_breakdown}
\end{figure}

\noindent\textbf{Results.} We report the manually checked cases in Figure~\ref{fig:error_percentage_breakdown}. The bar chart illustrates the percentage of each error that occurs in the incorrect sample. We observe three major errors that contributed to the failure of FoR-SALE.
The first error (23.33\% of all errors) is generation failure, which occurs when multiple compositions of the latent space overlap at the same pixel location, leading to poor-quality object images (example in the upper-left picture in Figure~\ref{fig:overall_fail}). This issue 
is closely related to the second source of error, object detection failure (18.33\% of all errors), which arises when two objects are positioned too closely. In such cases, the detector may fail to capture all objects or focus on incorrect regions of the image due to flawed generation.
The third and most significant source of error is incorrect orientation generation (28.33\% of all errors). 
As noted by~\citet{GenSpace}, even SOTA T2I models struggle to produce correct object orientations, highlighting a fundamental challenge that requires stronger 3D awareness in diffusion models. A similar challenge is observed for depth generation, where limited progress has been made in depth-editing actions. 
\begin{table}
    \centering
    \small
    \begin{tabular}{l c c} \\ \toprule  
    Baseline & Soft-Error & Hard-Error \\ \midrule
    SD 3.5-Large & $56.25$ & $73.20$\\
    FLUX.1 & $67.10$ & $78.71$\\ 
    GPT-4o (gpt-image-1) & $49.47$ & $72.35$ \\  \bottomrule
    \end{tabular}
    \caption{DiffEdit orientation-editing error rates after one round of FoR-SALE across three base generators.}\label{table:percentage_error_diff_edit}
\end{table} 
To further analyze DiffEdit’s orientation editing, we examine images from three base generators that underwent only orientation editing during one round of FoR-SALE. We report the error percentage of DiffEdit orientation edits in Table~\ref{table:percentage_error_diff_edit}. The soft-error metric allows more orientation labels to be considered correct (e.g., treating “left” as correct for backward-left and forward-left, and vice versa), whereas the hard-error metric requires the generated orientation to match the expected label exactly.

%% file: 99_prompt.tex
\section{Perspective Conversion Rules}\label{appendix:rules}

In this section, we present all manually crafted perspective-conversion rules used in the FoR Interpreter and their corresponding camera-perspective mappings.
The rules are categorized by the facing direction of the reference object.
Each facing direction is associated with exactly four conversion rules, corresponding to the four spatial relations considered in this work, i.e., left, right, front, and back.

\begin{enumerate}[label=\arabic*.]
    \item Facing toward the camera.
    
    \begin{enumerate}
        \item Left. If the object is facing toward the camera (front), then the left side of the object is on the right from the camera perspective.
        \item Right. If the object is facing toward the camera (front), then the right side of the object is on the left from the camera perspective.
        \item Front. If the object is facing toward the camera (front), then the front side of the object is in the front direction from the camera perspective.
        \item Back. If the object is facing toward the camera (front), then the back side of the object is in the back direction from the camera perspective.
    \end{enumerate}

    \item Facing forward-left.
    
    \begin{enumerate}
        \item Left. If the object is facing forward-left, then the left side of the object is on the right from the camera perspective.
        \item Right. If the object is facing forward-left, then the right side of the object is on the left from the camera perspective.
        \item Front. If the object is facing forward-left, then the front side of the object is in the front direction from the camera perspective.
        \item Back. If the object is facing forward-left, then the back side of the object is in the back direction from the camera perspective.
    \end{enumerate}

    \item Facing left.
    
    \begin{enumerate}
        \item Left. If the object is facing left, then the left side of the object is in the front direction from the camera perspective.
        \item Right. If the object is facing left, then the right side of the object is in the back direction from the camera perspective.
        \item Front. If the object is facing left, then the front side of the object is on the left from the camera perspective.
        \item Back. If the object is facing left, then the back side of the object is on the right from the camera perspective.
    \end{enumerate}

    \item Facing backward-left.
    
    \begin{enumerate}
        \item Left. If the object is facing backward-left, then the left side of the object is on the left from the camera perspective.
        \item Right. If the object is facing backward-left, then the right side of the object is on the right from the camera perspective.
        \item Front. If the object is facing backward-left, then the front side of the object is in the back direction from the camera perspective.
        \item Back. If the object is facing backward-left, then the back side of the object is in the front direction from the camera perspective.
    \end{enumerate}

    \item Facing away from the camera.
    
    \begin{enumerate}
        \item Left. If the object is facing away from the camera (back), then the left side of the object is on the left from the camera perspective.
        \item Right. If the object is facing away from the camera (back), then the right side of the object is on the right from the camera perspective.
        \item Front. If the object is facing away from the camera (back), then the front side of the object is in the back direction from the camera perspective.
        \item Back. If the object is facing away from the camera (back), then the back side of the object is in the front direction from the camera perspective.
    \end{enumerate}

    \item Facing backward-right.
    
    \begin{enumerate}
        \item Left. If the object is facing backward-right, then the left side of the object is on the left from the camera perspective.
        \item Right. If the object is facing backward-right, then the right side of the object is on the right from the camera perspective.
        \item Front. If the object is facing backward-right, then the front side of the object is in the back direction from the camera perspective.
        \item Back. If the object is facing backward-right, then the back side of the object is in the front direction from the camera perspective.
    \end{enumerate}

    \item Facing right.
    
    \begin{enumerate}
        \item Left. If the object is facing right, then the left side of the object is in the back direction from the camera perspective.
        \item Right. If the object is facing right, then the right side of the object is in the front direction from the camera perspective.
        \item Front. If the object is facing right, then the front side of the object is on the right from the camera perspective.
        \item Back. If the object is facing right, then the back side of the object is on the left from the camera perspective.
    \end{enumerate}

    \item Facing forward-right.
    
    \begin{enumerate}
        \item Left. If the object is facing forward-right, then the left side of the object is on the right from the camera perspective.
        \item Right. If the object is facing forward-right, then the right side of the object is on the left from the camera perspective.
        \item Front. If the object is facing forward-right, then the front side of the object is in the front direction from the camera perspective.
        \item Back. If the object is facing forward-right, then the back side of the object is in the back direction from the camera perspective.
    \end{enumerate}
    
\end{enumerate}

\section{Human Alignment Score}\label{sec:human_alignmet_score}

\begin{lstlisting}[caption={Instruction for collecting human results on evaluating generated images.}, label={lst:human_instruction}]
"""
Instruction: 
You will be presented with an image generated by a generative model along with its corresponding input prompt. Your task is to determine whether the generated image aligns with the prompt. If the image is aligned, assign a score of 1; otherwise, assign a score of 0.
"""
\end{lstlisting}

\subsection{Experimental setting}
We provide a set of 200 images—sampled from the initial FLUX.1 generations and the first round of FoR-SALE—to a human annotator, compensated as a research assistant. This selection ensures diversity in the generated outputs and maintains independence from our specific generation framework. The annotator follows the instructions in Listing~\ref{lst:human_instruction} to judge whether each image aligns with the corresponding input description. Each sample is labeled as either aligned (1) or not aligned (0), and we collect these binary scores for all images.

\section{Prompt Specifications}\label{sec:LLM_prompt}

In this section, we provide the prompts used by the LLM components throughout the entire experiment. The listings show the static prompt templates. Task-specific inputs are appended programmatically at runtime. For example, \{conversion\_rules\} and \{remaining\_examples\} denote dynamically inserted content; the 32 manually defined rules are provided in Appendix~\ref{appendix:rules}.

\begin{lstlisting}[caption={Prompt for generating a layout for GLIGEN.}, label={lst:LLM_GLIGEN}]
Your task is to generate the bounding boxes of objects mentioned in the caption, along with the direction that the objects are facing.
The image is size 512x512.
The bounding box should be in the format of [x, y, width, height], with each value from 0 to 1.
The direction that the object is facing should be one of these options: [front, back, left, right].
Please consider the frame of reference of the caption and the direction of the reference object.
The answer should be in the form:
Reasoning: Explanation
Layout: Layout

The example of layout is [('cat', [0.1, 0.3, 0.5, 0.4], 'right'), ('cow', [0.6, 0.5, 0.3, 0.4], 'right')]

User prompt: {prompt}
Current Objects: {current_objects}
\end{lstlisting}

\begin{lstlisting}[caption={Prompt for LLM Parser.}, label={lst:LLM_parser}]
# Your Role: Excellent Parser

## Objective: Analyze scene descriptions to identify objects and their attributes.

## Process Steps
1. Read the user prompt (scene description).
2. Identify all objects mentioned with quantities.
3. Extract attributes of each object (color, size, material, etc.).
4. Ignore facing attribute (facing to left, facing to right, facing forward).
5. If the description mentions objects that shouldn't be in the image, take note of the negation part.
6. Explain your understanding (reasoning) and then format your result (answer/negation) as shown in the examples.
7. Importance of Extracting Attributes: Attributes provide specific details about the objects. This helps differentiate between similar objects and gives a clearer understanding of the scene.

## Examples

- Example 1
  User prompt: A brown horse is beneath a black dog. Another orange cat is beneath a brown horse.
  Reasoning: The description talks about three objects: a brown horse, a black dog, and an orange cat. We report the color attribute thoroughly. No specified negation terms. No background is mentioned and thus fill in the default one.
  Objects: [('horse', ['brown']), ('dog', ['black']), ('cat', ['orange'])]
  Background: A realistic image
  Negation: 

- Example 2
  User prompt: There's a white car and a yellow airplane in a garage. They're in front of two dogs and behind a cat. The car is small. Another yellow car is outside the garage.
  Reasoning: The scene has two cars, one airplane, two dogs, and a cat. The car and airplane have colors. The first car also has a size. No specified negation terms. The background is a garage.
  Objects: [('car', ['white and small', 'yellow']), ('airplane', ['yellow']), ('dog', [None, None]), ('cat', [None])]
  Background: A realistic image in a garage
  Negation: 

- Example 3
  User prompt: A car and a dog are on top of an airplane and below a red chair. There's another dog sitting on the mentioned chair.
  Reasoning: Five object instances are described: one car, one airplane, two dogs, and one red chair. No specified negation terms. No background is mentioned, and thus fill in the default one.
  Objects: [('car', [None]), ('airplane', [None]), ('dog', [None, None]), ('chair', ['red'])]
  Background: A realistic image
  Negation: 

- Example 4
  User prompt: An oil painting at the beach of a blue bicycle to the left of a bench and to the right of a palm tree with five seagulls in the sky.
  Reasoning: Here, there are five seagulls, one blue bicycle, one palm tree, and one bench. No specified negation terms. The background is an oil painting at the beach.
  Objects: [('bicycle', ['blue']), ('palm tree', [None]), ('seagull', [None, None, None, None, None]), ('bench', [None])]
  Background: An oil painting at the beach
  Negation: 

- Example 5
  User prompt: An animated-style image of a scene without backpacks.
  Reasoning: The description clearly states no backpacks, so this must be acknowledged. The user provides the negative prompt of backpacks. The background is an animated-style image.
  Objects: [('backpacks', [None])]
  Background: An animated-style image
  Negation: backpacks

- Example 6
  User Prompt: Make the dog a sleeping dog and remove all shadows in an image of a grassland.
  Reasoning: The user prompt specifies a sleeping dog and requests that all shadows be removed. The background is a realistic image of a grassland.
  Objects: [('dog', ['sleeping']), ('shadow', [None])]
  Background: A realistic image of a grassland
  Negation: shadows

- Example 7
  User Prompt: A fire hydrant is behind a cat relative to the observer. The cat is facing away from the observer.
  Reasoning: Two objects are described: one fire hydrant and one cat. The facing direction is ignored as instructed. No specified negation terms. No background is mentioned, and thus fill in the default one.
  Objects: [('fire hydrant', [None]), ('cat', [None])]
  Background: A realistic image
  Negation: 

Your Current Task: Follow the steps closely and accurately identify objects based on the given prompt. Ensure adherence to the above output format.

User prompt: {prompt}
Objects: {current_objects}
Background: {background}
Negation: {negation}
\end{lstlisting}

\begin{lstlisting}[caption={Prompt for FoR Interpreter.}, label={lst:FoR_Interpreter}]
# Your Role: Expert on spatial relations in multiple perspectives

## Objective: Interpret the prompt and convert the spatial relation into the camera's perspective

## Image and Object Specification
1. Image Coordinates: Define square images with top-left at [0, 0] and bottom-right at [1, 1].
2. Four pieces of information are given for each object in order: object name, bounding box, depth, and facing direction.
3. Object Format: (object, box, depth, facing direction)
4. Box Format: [Top-left x, Top-left y, Width, Height]
5. Depth: Define depth of the object from furthest at 0 and nearest at 1.
6. Facing Direction: An orientation of the object relative to the camera, which can be None, left, forward-left, backward-left, right, forward-right, backward-right, front (facing forward or facing toward), or back (facing backward or facing away).

## Key Guidelines
1. Perspective Identification: Carefully consider the perspective of the spatial relation presented in the prompt.
2. Object facing direction: Carefully consider the facing orientation presented in the prompt first, before considering the facing orientation from the object specification.
3. Assume the camera, observer, and I (me) are the same thing and have the same view (perspective).
4. Look at the example closely to see how the conversion needs to be made.

<RULES>
{conversion_rules}
</RULES>

## Process Steps
1. Read and understand the user prompt (scene description).
2. Identify the perspective of the spatial relation presented in the given prompt.
3. Check whether the facing direction is provided in the prompt.
4. If not, check the facing direction presented in the object specification.
5. Explain your understanding (reasoning) and then convert the perspective into the camera's perspective.
6. If there is no specification of perspective, assume the camera perspective for minimal editing of the given prompt.
7. Do not modify any other part of the prompt except for spatial relation(s).
8. Do not update the object, only modify the prompt.

## Examples

- Example 1
  User prompt: a backpack on the right of a car from the car's perspective and a car on the left
  Current Objects: [('backpack #1', [0.302, 0.293, 0.335, 0.194], 0.63, None), ('car #1', [0.027, 0.324, 0.246, 0.160], 0.25, 'left')]
  Reasoning: There are two spatial relations presented in the prompt. The first one specifies a backpack on the right of a car from "the car's perspective." There is no specific facing direction of the car presented in the prompt. Therefore, consider the car's facing direction in the object's current state ('left'). The car is facing to the left of the photo. Therefore, the right of the car from the car's perspective corresponds to behind the car from the camera's perspective. The second spatial relation is that the car is on the left. This does not specify the perspective. Therefore, assume a camera perspective for this one and make no update to the second spatial relation.
  Updated prompt: a backpack behind a car from the camera's perspective and a car on the left

- Example 2
  User prompt: a cat is on the left and the cup is on the right of the cat from the cat's view
  Current Objects: [('cat #1', [0.169, 0.563, 0.323, 0.291], 0.901, 'right'), ('cup #1', [0.59, 0.186, 0.408, 0.814], 0.732, None)]
  Reasoning: There are two spatial relations presented in the prompt. The first spatial relation is a cat on the left. The prompt does not specify the perspective. Therefore, assume a camera perspective for this one and make no update. The second one specifies that the cup is on the right of the cat from "the cat's view." There is no specific facing direction of the cat presented in the prompt. Therefore, consider the cat's facing direction in the object's current state ('right'). The cat is facing to the right of the photo. Therefore, the right of the cat from the cat's perspective corresponds to in front of the cat from the camera's perspective.
  Updated prompt: a cat is on the left and the cup is in front of the cat from the camera's view

- Example 3
  User prompt: A cow is in front of a sheep from the camera angle. The sheep is facing right relative to the camera.
  Current Objects: [('cow #1', [0.354, 0.365, 0.285, 0.385], 0.41, None), ('sheep #1', [0.608, 0.120, 0.285, 0.200], 0.82, 'right')]
  Reasoning: There is only one spatial relation presented in the prompt. The prompt specifies that a cow is in front of a sheep from the "camera angle." This spatial relation is from the camera's perspective. Therefore, there is no need for change.
  Updated prompt: A cow is in front of a sheep from the camera angle. The sheep is facing right relative to the camera.

- Example 4
  User prompt: A fire hydrant is back of a sheep from the sheep's perspective. The sheep is facing away from the camera.
  Current Objects: [('fire hydrant #1', [0.113, 0.365, 0.251, 0.251], 0.64, None), ('sheep #1', [0.608, 0.120, 0.251, 0.251], 0.52, 'back')]
  Reasoning: There is only one spatial relation presented in the prompt. The prompt specifies that a fire hydrant is behind a sheep from "the sheep's perspective." The prompt also specifies that the sheep is facing away (back) from the camera. Therefore, the back of the sheep corresponds to the front direction of the camera. The updated spatial relation is that the fire hydrant is in front of the sheep from the camera's perspective.
  Updated prompt: A fire hydrant is in front of a sheep from the camera's perspective. The sheep is facing away from the camera.

- Example 5
  User prompt: A deer is to the left of a car from the car's perspective. The car is facing away from the camera.
  Current Objects: [('deer #1', [0.454, 0.165, 0.285, 0.385], 0.42, None), ('car #1', [0.608, 0.620, 0.285, 0.200], 0.83, 'back')]
  Reasoning: There is only one spatial relation presented in the prompt. The prompt specifies that a deer is to the left of a car from "the car's perspective." The prompt also specifies that the car is facing away (back) from the camera. Therefore, the left side of the car corresponds to the left direction of the camera. The updated spatial relation is that the deer is to the left of the car from the camera's perspective.
  Updated prompt: A deer is to the left of a car from the camera's perspective. The car is facing away from the camera.

- Example 6
  User prompt: A cow is to the right of a horse from the horse's perspective. The horse is facing toward relative to the camera.
  Current Objects: [('cow #1', [0.113, 0.365, 0.352, 0.352], 0.83, None), ('horse #1', [0.608, 0.120, 0.352, 0.352], 0.25, 'front')]
  Reasoning: There is only one spatial relation presented in the prompt. The prompt specifies that a cow is to the right of a horse from "the horse's perspective." The prompt also specifies that the horse is facing toward (front) the camera. Therefore, the right side of the horse corresponds to the left direction of the camera. The updated spatial relation is that the cow is to the left of the horse from the camera's perspective.
  Updated prompt: A cow is to the left of a horse from the camera's perspective. The horse is facing toward relative to the camera.

- Example 7
  User prompt: A deer is in front of a sheep from the sheep's perspective. The sheep is facing toward relative to the camera.
  Current Objects: [('deer #1', [0.454, 0.365, 0.285, 0.385], 0.64, None), ('sheep #1', [0.608, 0.120, 0.285, 0.200], 0.32, 'front')]
  Reasoning: There is only one spatial relation presented in the prompt. The prompt specifies that a deer is in front of a sheep from "the sheep's perspective." The prompt also specifies that the sheep is facing toward (front) the camera. Therefore, the front of the sheep corresponds to the front direction of the camera. The updated spatial relation is that the deer is in front of the sheep from the camera's perspective.
  Updated prompt: A deer is in front of a sheep from the camera's perspective. The sheep is facing toward relative to the camera.

- Example 8
  User prompt: A deer is in front of a dog from the dog's perspective. The dog is facing right relative to the camera.
  Current Objects: [('deer #1', [0.186, 0.592, 0.449, 0.408], 0.45, 'front'), ('dog #1', [0.376, 0.194, 0.624, 0.502], 0.53, 'right')]
  Reasoning: There is only one spatial relation presented in the prompt. The prompt specifies that a deer is in front of a dog from "the dog's perspective." The prompt also specifies that the dog is facing to the right relative to the camera. Therefore, the front of the dog corresponds to the right direction of the camera. The updated spatial relation is that the deer is to the right of the dog from the camera's perspective.
  Updated prompt: A deer is to the right of a dog from the camera's perspective. The dog is facing right relative to the camera.

- Example 9
  User prompt: A deer is to the right of a car from the car's perspective. The car is facing away from the camera.
  Current Objects: [('deer #1', [0.454, 0.165, 0.285, 0.385], 0.42, None), ('car #1', [0.608, 0.620, 0.285, 0.200], 0.83, 'back')]
  Reasoning: There is only one spatial relation presented in the prompt. The prompt specifies that a deer is to the right of a car from "the car's perspective." The prompt also specifies that the car is facing away (back) from the camera. Therefore, the right side of the car corresponds to the right direction of the camera. The updated spatial relation is that the deer is to the right of the car from the camera's perspective.
  Updated prompt: A deer is to the right of a car from the camera's perspective. The car is facing away from the camera.

Your Current Task: Follow the steps closely and accurately convert all presented spatial relations in the given prompt into the camera's perspective. Ensure adherence to the above output format.

User prompt: {prompt}
Current Objects: {current_objects}
\end{lstlisting}

\begin{lstlisting}[caption={Prompt for Layout Interpreter.}, label={lst:Layout_Interpreter}]
# Your Role: Expert Bounding Box Adjuster

## Objective: Manipulate bounding boxes in square images according to the user prompt while maintaining visual accuracy.

## Object Specifications and Manipulations
1. Image Coordinates: Define square images with top-left at [0, 0] and bottom-right at [1, 1].
2. Object Format: (object, box, depth, orientation)
3. Box Format: [Top-left x, Top-left y, Width, Height]
4. Depth: Define the depth of the object, with furthest at 0 and nearest at 1.
5. Orientation Format: An orientation of the object, which can be None, 'left', 'right', 'front', 'back', 'forward-left', 'forward-right', 'backward-left', or 'backward-right'.
6. Operations: Include addition, deletion, repositioning, attribute modification, and depth modification.

## Key Guidelines
1. Alignment: Follow the user's prompt, keeping the specified object count and attributes. Deem it incorrect if the described object lacks specified attributes.
2. Boundary Adherence: Keep bounding box coordinates within [0, 1].
3. Depth Adherence: Keep each object's depth within[0, 1].
4. Orientation Adherence: An orientation must change depending on the prompt. If nothing is specified in the prompt, do not change the orientation of the object.
5. Minimal Modifications: Change bounding boxes or depth only if they do not match the user's prompt (i.e., do not modify matched objects).
6. Overlap Reduction: Minimize intersections in new boxes and remove the smallest, least overlapping objects when an object needs to be removed.
7. Left/Right Relation: Compare the horizontal center of each bounding box, where center_x = x + width / 2. Object A is left of object B if center_x(A) < center_x(B), and right of object B if center_x(A) > center_x(B).
8. Above/Below Relation: Compare the vertical center of each bounding box, where center_y = y + height / 2. Object A is above object B if center_y(A) < center_y(B), and below object B if center_y(A) > center_y(B).
9. Front/Behind Relation: Compare object depth. Object A is in front of object B if depth(A) > depth(B), and behind object B if depth(A) < depth(B).

## Process Steps
1. Interpret prompts: Read and understand the user's prompt.
2. Implement Changes: Review and adjust current bounding boxes to meet user specifications.
3. Explain Adjustments: Justify the reasons behind each alteration and ensure every adjustment abides by the key guidelines.
4. Output the Result: Present the reasoning first, followed by the updated objects section, which should include a list of bounding boxes in Python format.

## Examples

- Example 1
  User prompt: A realistic image of landscape scene depicting a green car parking on the left of a blue truck, with a red air balloon and a bird in the sky
  Current Objects: [('green car #1', [0.027, 0.365, 0.275, 0.207], 0.6, None), ('blue truck #1', [0.350, 0.368, 0.272, 0.208], 0.7, None), ('red air balloon #1', [0.086, 0.010, 0.189, 0.176], 0.4, None)]
  Reasoning: The green car is already to the left of the blue truck, and the red air balloon is already present. The prompt also specifies a bird in the sky, but no bird is present in the current objects. Therefore, add a bird in an upper region of the image while keeping all existing matched objects unchanged and ensuring all coordinates and dimensions remain within [0, 1].
  Updated Objects: [('green car #1', [0.027, 0.365, 0.275, 0.207], 0.6, None), ('blue truck #1', [0.350, 0.368, 0.272, 0.208], 0.7, None), ('red air balloon #1', [0.086, 0.010, 0.189, 0.176], 0.4, None), ('bird #1', [0.385, 0.054, 0.186, 0.130], 0.3, None)]

- Example 2
  User prompt: A realistic image of landscape scene depicting a green car parking on the right of a blue truck, with a red air balloon and a bird in the sky
  Current Objects: [('green car #1', [0.027, 0.365, 0.275, 0.207], 0.79, 'left'), ('blue truck #1', [0.350, 0.369, 0.272, 0.208], 0.68, 'right'), ('red air balloon #1', [0.086, 0.010, 0.189, 0.176], 0.15, None)]
  Reasoning: The green car is currently to the left of the blue truck, but the prompt requires the green car to be to the right of the blue truck. Therefore, reposition the car and truck so that their left-right relation matches the prompt. The prompt also specifies a bird in the sky, but no bird is present in the current objects, so add one in an upper image region while minimizing overlap. Keep all other matched objects unchanged.
  Updated Objects: [('green car #1', [0.350, 0.369, 0.275, 0.207], 0.79, 'left'), ('blue truck #1', [0.027, 0.365, 0.272, 0.208], 0.68, 'right'), ('red air balloon #1', [0.086, 0.010, 0.189, 0.176], 0.15, None), ('bird #1', [0.485, 0.054, 0.186, 0.130], 0.15, 'front')]

- Example 3
  User prompt: An oil painting of a pink dolphin jumping on the left of a steam boat on the sea
  Current Objects: [('steam boat #1', [0.302, 0.293, 0.335, 0.194], 0.76, 'front'), ('pink dolphin #1', [0.027, 0.324, 0.246, 0.160], 0.23, 'left'), ('blue dolphin #1', [0.158, 0.454, 0.376, 0.290], 0.26, 'right')]
  Reasoning: The prompt mentions only one dolphin, but two dolphins are present. The pink dolphin already matches the prompt, so remove the extra blue dolphin while keeping the matched objects unchanged.
  Updated Objects: [('steam boat #1', [0.302, 0.293, 0.335, 0.194], 0.76, 'front'), ('pink dolphin #1', [0.027, 0.324, 0.246, 0.160], 0.23, 'left')]

- Example 4
  User prompt: An oil painting of a pink dolphin jumping on the left of a steam boat on the sea
  Current Objects: [('steam boat #1', [0.302, 0.293, 0.335, 0.194], 0.76, 'front'), ('dolphin #1', [0.027, 0.324, 0.246, 0.160], 0.23, 'left')]
  Reasoning: The prompt specifies a pink dolphin, but the current object is a generic dolphin without the required pink attribute. Therefore, modify the dolphin's attribute while keeping its position, depth, and orientation unchanged.
  Updated Objects: [('steam boat #1', [0.302, 0.293, 0.335, 0.194], 0.76, 'front'), ('pink dolphin #1', [0.027, 0.324, 0.246, 0.160], 0.23, 'left')]

- Example 5
  User prompt: a backpack on the right of a car from the car's perspective and a car on the left
  Current Objects: [('backpack #1', [0.302, 0.293, 0.335, 0.194], 0.63, None), ('car #1', [0.027, 0.324, 0.246, 0.160], 0.25, 'left')]
  Reasoning: The prompt specifies that a backpack is on the right of a car from the car's perspective. There is no specific orientation of the car in the prompt, so use the current car orientation, which is 'left'. When the car faces left relative to the camera, its intrinsic right corresponds to behind the car from the camera's perspective. Therefore, the backpack should have a lower depth than the car. The current backpack depth (0.63) is higher than the car depth (0.25), so the depth relation does not match the prompt. Swap the depths of the car and backpack while keeping their bounding boxes and orientations unchanged.
  Updated Objects: [('backpack #1', [0.302, 0.293, 0.335, 0.194], 0.25, None), ('car #1', [0.027, 0.324, 0.246, 0.160], 0.63, 'left')]

- Example 6
  User prompt: a cat is on the left and the cup is on the right of the cat from the cat's view
  Current Objects: [('cat #1', [0.169, 0.563, 0.323, 0.291], 0.901, 'right'), ('cup #1', [0.59, 0.186, 0.408, 0.814], 0.732, None)]
  Reasoning: The prompt specifies that the cat is on the left, which is already satisfied. The prompt also specifies that the cup is on the right of the cat from the cat's view. The prompt does not specify the cat's orientation, so use the current orientation, which is 'right'. When the cat faces right relative to the camera, its intrinsic right corresponds to in front of the cat from the camera's perspective. Therefore, the cup's depth must be greater than the cat's depth. The cat currently has depth 0.901 while the cup has depth 0.732. To make the minimal modification, keep the cat unchanged and increase the cup's depth to a value greater than 0.901.
  Updated Objects: [('cat #1', [0.169, 0.563, 0.323, 0.291], 0.901, 'right'), ('cup #1', [0.59, 0.186, 0.408, 0.814], 0.95, None)]

- Example 7
  User prompt: A cow is in front of a sheep from the camera angle. The sheep is facing right relative to the camera.
  Current Objects: [('cow #1', [0.354, 0.365, 0.285, 0.385], 0.41, None), ('sheep #1', [0.608, 0.120, 0.285, 0.200], 0.82, 'right')]
  Reasoning: The prompt specifies that a cow is in front of a sheep from the camera angle. Therefore, the cow's depth should be greater than the sheep's depth. However, the cow depth (0.41) is lower than the sheep depth (0.82), which does not match the prompt. Swap the depths of the cow and sheep while keeping all other properties unchanged.
  Updated Objects: [('cow #1', [0.354, 0.365, 0.285, 0.385], 0.82, None), ('sheep #1', [0.608, 0.120, 0.285, 0.200], 0.41, 'right')]

- Example 8
  User prompt: A fire hydrant is back of a sheep from the sheep's perspective. The sheep is facing left relative to the camera.
  Current Objects: [('fire hydrant #1', [0.113, 0.365, 0.251, 0.251], 0.64, None), ('sheep #1', [0.608, 0.120, 0.251, 0.251], 0.52, 'left')]
  Reasoning: The prompt specifies that a fire hydrant is behind a sheep from the sheep's perspective. Since the sheep is facing left relative to the camera, the sheep's intrinsic back corresponds to the camera's right direction. Therefore, the fire hydrant should be to the right of the sheep. The fire hydrant is currently to the left of the sheep, so reposition only the fire hydrant horizontally to the right of the sheep while preserving its size, depth, orientation, and approximate vertical position.
  Updated Objects: [('fire hydrant #1', [0.700, 0.365, 0.251, 0.251], 0.64, None), ('sheep #1', [0.608, 0.120, 0.251, 0.251], 0.52, 'left')]

- Example 9
  User prompt: A cow is to the left of a horse from the horse's perspective. The horse is facing right relative to the camera.
  Current Objects: [('cow #1', [0.113, 0.365, 0.352, 0.352], 0.83, None), ('horse #1', [0.608, 0.120, 0.352, 0.352], 0.25, 'right')]
  Reasoning: The prompt specifies that a cow is to the left of a horse from the horse's perspective. Since the horse is facing right relative to the camera, the horse's intrinsic left corresponds to behind the horse from the camera's perspective. Therefore, the cow's depth should be lower than the horse's depth. However, the cow depth (0.83) is higher than the horse depth (0.25), which does not match the prompt. Swap the depths of the cow and horse while keeping all other properties unchanged.
  Updated Objects: [('cow #1', [0.113, 0.365, 0.352, 0.352], 0.25, None), ('horse #1', [0.608, 0.120, 0.352, 0.352], 0.83, 'right')]

- Example 10
  User prompt: A deer is in front of a car from the car's perspective. The car is facing toward the camera.
  Current Objects: [('deer #1', [0.454, 0.365, 0.285, 0.385], 0.64, None), ('car #1', [0.608, 0.120, 0.285, 0.200], 0.32, 'left')]
  Reasoning: The prompt specifies that a deer is in front of a car from the car's perspective. Since the car is facing toward the camera, the car's intrinsic front corresponds to in front of the car from the camera's perspective. The deer depth (0.64) is higher than the car depth (0.32), which already matches the required spatial relation. However, the current orientation of the car is 'left', while the prompt explicitly specifies that the car is facing toward the camera. Therefore, change only the car's orientation to 'front'.
  Updated Objects: [('deer #1', [0.454, 0.365, 0.285, 0.385], 0.64, None), ('car #1', [0.608, 0.120, 0.285, 0.200], 0.32, 'front')]

- Example 11
  User prompt: A deer is in front of a car from the car's perspective. The car is facing away from the camera.
  Current Objects: [('deer #1', [0.454, 0.165, 0.285, 0.385], 0.42, None), ('car #1', [0.608, 0.620, 0.285, 0.200], 0.83, 'back')]
  Reasoning: The prompt specifies that a deer is in front of a car from the car's perspective. Since the car is facing away from the camera, the car's intrinsic front corresponds to behind the car from the camera's perspective. Therefore, the deer should have a lower depth than the car. The deer depth (0.42) is lower than the car depth (0.83), so the current image already matches the prompt and requires no modification.
  Updated Objects: [('deer #1', [0.454, 0.165, 0.285, 0.385], 0.42, None), ('car #1', [0.608, 0.620, 0.285, 0.200], 0.83, 'back')]
  
- Example 12
  User prompt: A realistic photo of a scene with a brown bowl on the right and a gray dog on the left
  Current Objects: [('gray dog #1', [0.186, 0.592, 0.449, 0.408], 0.45, 'front'), ('brown bowl #1', [0.376, 0.194, 0.624, 0.502], 0.53, None)]
  Reasoning: The horizontal center of the gray dog is 0.186 + 0.449 / 2 = 0.4105, while the horizontal center of the brown bowl is 0.376 + 0.624 / 2 = 0.688. Since the gray dog's center is to the left of the brown bowl's center, the required left-right relation is already satisfied. No modification is needed.
  Updated Objects: [('gray dog #1', [0.186, 0.592, 0.449, 0.408], 0.45, 'front'), ('brown bowl #1', [0.376, 0.194, 0.624, 0.502], 0.53, None)]

Your Current Task: Carefully follow the provided guidelines and steps to adjust bounding boxes in accordance with the user's prompt. Ensure adherence to the above output format.

User prompt: {prompt}
Current Objects: {current_objects}
\end{lstlisting}

\begin{lstlisting}[caption={Prompt for Extract Cognitive Map before Generating Image.}, label={lst:Cognative_map}]
Please also consider performing the cognitive map before generating the image.

Cognitive Map's objective is to identify specific objects from the context,
understand the spatial arrangement of the scene, and estimate the center point of each object, assuming the entire scene is represented by a 10x10 grid.

[Rule]
1. Estimate the center location of each instance within the provided categories, assuming the entire scene is represented by a 10x10 grid.
2. If a category contains multiple instances, include all of them.
3. Each object's estimated location should accurately reflect its real position in the scene, preserving the relative spatial relationships among all objects.

Still, you need to generate the real image, not a cognitive map.
\end{lstlisting}

\begin{lstlisting}[caption={Prompt for LLM to humanize the given context.}, label={lst:humanized_prompt}]
Instruction:
You rewrite spatial-reasoning prompts to sound more natural and varied while preserving all spatial facts.

Rules:
- Preserve every object, relation, perspective, and facing direction.
- Do not add new objects, remove objects, or change object order.
- Do not change the meaning of relative, intrinsic, camera, image, or object-perspective relations.
- Keep the prompt concise and image-generation friendly.
- Return only valid JSON.

Rewrite the initial prompt into a more natural human-written prompt.

Return this JSON object:
{{
  "humanized_prompt": "string",
  "preserved": true,
  "notes": "short explanation of what changed"
}}

Initial prompt:
{prompt}

Structured relation info:
{relation_info}
\end{lstlisting}

\begin{lstlisting}[caption={Prompt for classifying frame of reference of given context used in the FoR Interpreter routing experiment.}, label={lst:routing_ex}]
Your Role: Expert on spatial relation frame-of-reference classification

## Objective
Interpret the prompt and decide whether the spatial relation is intrinsic or relative.

## Key Guidelines
1. Perspective Identification: Carefully consider the perspective of the spatial relation presented in the prompt.
2. Assume the camera, observer, and I (me) are the same thing and have the same view (perspective).
3. If the prompt refers to an object's own perspective, classify it as intrinsic.
4. If the prompt refers to the camera, observer, or my perspective, classify it as relative.
5. If there is at least one intrinsic spatial relation, the sentence is intrinsic.
6. Return only the requested tag format and do not add extra labels.

## Process Steps
1. Read and understand the user prompt (scene description).
2. Identify the perspective of each spatial relation presented in the given prompt.
3. Decide whether each relation is intrinsic or relative.
4. If there is at least one intrinsic relation, classify the overall prompt as intrinsic; otherwise, classify it as relative.
5. Explain your reasoning and then provide the final category.
6. Do not modify the prompt or the object specification.

## Examples

- Example 1
  User prompt: A fire hydrant is back of a cat relative to observer. The cat is facing away from the observer.
  Reasoning: The spatial relation is relative to the observer, so this is relative.
  Output: <reasoning> The spatial relation is relative to the observer, so this is relative. </reasoning> <answer> relative </answer>

- Example 2
  User prompt: A bird is inside the car and left of the car from the car's perspective.
  Reasoning: The left relation is described from the car's own perspective, so this is intrinsic.
  Output: <reasoning> The left relation is described from the car's own perspective, so this is intrinsic. </reasoning> <answer> intrinsic </answer>

- Example 3
  User prompt: A phone is on the left of a tablet from my perspective.
  Reasoning: The relation is from an outside observer's perspective, so this is relative.
  Output: <reasoning> The relation is from an outside observer's perspective, so this is relative. </reasoning> <answer> relative </answer>

Your Current Task: Follow the steps closely and accurately classify the spatial relation in the given prompt. Return the result in the exact format: <reasoning> Reasoning </reasoning> <answer> intrinsic/relative </answer>

User prompt: {prompt}
\end{lstlisting}

\begin{lstlisting}[caption={Prompt for generated rules used in FoR Interpreter.}, label={lst:generate_rule}]
# Your Role: Expert Perspective Conversion Rule Generator

## Objective
Generate deterministic perspective-conversion rules for spatial reasoning.
The rules should convert object-centric spatial relations into equivalent camera-centric spatial relations.

## Perspective Conversion Specifications
1. Camera-centric perspective: The spatial relation is interpreted from the image/camera view.
2. Object-centric perspective: The spatial relation is interpreted from the relatum/object's own facing direction.
3. Relatum: The reference object whose perspective is used.
4. Target: The object being located relative to the relatum.
5. Conversion rule: A mapping from an object-centric relation and a relatum facing direction to a camera-centric relation.

## Key Guidelines
1. Relation Alignment: Follow the given relation set exactly. Do not invent new relations.
2. Facing Direction Alignment: Follow the given facing direction set exactly. Do not invent new facing directions.
3. Deterministic Mapping: Each relation-facing pair must map to exactly one camera-centric relation.
4. Complete Coverage: Generate one rule for every combination of relation and facing direction.
5. Extractability: The output must be directly parseable using Python with json.loads().
6. Minimal Output: Return only the structured JSON object.
7. Use the example rules below as valid demonstrations of the expected mapping style and wording.

## Process Steps
1. Interpret Inputs: Read the given relation set and facing direction set.
2. Generate Mappings: For each relation-facing pair, infer the equivalent camera-centric relation.
3. Explain Each Rule: Write a concise natural-language rule explaining the conversion.
4. Output the Result: Present the final rules in the required JSON format.

## Valid Example Rules

The following grouped dictionary is a valid example of perspective-conversion rules.
Use it as the reference for both mapping behavior and wording style.

Example Relations:
["left", "right", "front", "back"]

Example Facing Directions:
["front", "forward-left", "left", "backward-left", "back", "backward-right", "right", "forward-right"]

Example Rule Dictionary:
{
    "front": [
        "If the object is facing toward the camera (front), then the left side of the object is on the right from the camera perspective.",
        "If the object is facing toward the camera (front), then the right side of the object is on the left from the camera perspective.",
        "If the object is facing toward the camera (front), then the front side of the object is in the front direction from the camera perspective.",
        "If the object is facing toward the camera (front), then the back side of the object is in the back direction from the camera perspective."
    ],
{remaining_examples}
}

Important:
- This example is provided in grouped dictionary format only as a demonstration.
- Your output must NOT use the grouped dictionary format.
- Your output must use the structured JSON schema with the top-level key "rules".
- Each string in the example corresponds to one structured rule object.

## Output Format
Return ONLY valid JSON.
Do not include markdown.
Do not include comments.
Do not include trailing commas.
Do not include any explanation outside the JSON object.

The JSON must follow exactly this schema:

{
  "rules": [
    {
      "object_perspective_relation": "<relation>",
      "relatum_facing_direction": "<facing_direction>",
      "camera_perspective_relation": "<converted_relation>",
      "rule_text": "If the object is facing <facing_description>, then the <relation> side of the object is <converted_relation_description> from the camera perspective."
    }
  ]
}

## Constraints
- Include exactly one rule for every relation-facing_direction pair.
- Use only relations from the provided relation set.
- Use only facing directions from the provided facing direction set.
- The value of "camera_perspective_relation" must also be one of the provided relations.
- Do not invent new relation names.
- Do not invent new facing directions.
- Do not omit any relation-facing pair.
- Return JSON only.

## Current Task
Relations:
{relations}

Facing directions:
{facing_directions}

Carefully follow the provided guidelines and steps to generate deterministic perspective-conversion rules.
Return only the JSON object.
\end{lstlisting}

%% file: colm2026_conference.bib
@inproceedings{FoREST,
    title = "{F}o{REST}: Frame of Reference Evaluation in Spatial Reasoning Tasks",
    author = "Premsri, Tanawan and Kordjamshidi, Parisa",
    booktitle = "Proceedings of the 2025 Conference on Empirical Methods in Natural Language Processing",
    year = "2025",
    publisher = "Association for Computational Linguistics",
    pages = "34977--35003",
    url = "https://aclanthology.org/2025.emnlp-main.1772/"
}

@inproceedings{
COMFORT,
title={Do Vision-Language Models Represent Space and How? Evaluating Spatial Frame of Reference under Ambiguities},
author={Zheyuan Zhang and Fengyuan Hu and Jayjun Lee and Freda Shi and Parisa Kordjamshidi and Joyce Chai and Ziqiao Ma},
booktitle={The Thirteenth International Conference on Learning Representations},
year={2025},
url={https://openreview.net/forum?id=84pDoCD4lH}
}

@inproceedings{GenSpace,
  title={GenSpace: Benchmarking Spatially-Aware Image Generation},
  author={Zehan Wang and Jiayang Xu and Ziang Zhang and Tianyu Pang and Chao Du and Hengshuang Zhao and Zhou Zhao},
  booktitle={Advances in Neural Information Processing Systems},
  year={2025},
  note={Datasets and Benchmarks Track}
}

@InProceedings{Wu_SLD,
    author    = {Wu, Tsung-Han and Lian, Long and Gonzalez, Joseph E. and Li, Boyi and Darrell, Trevor},
    title     = {Self-correcting LLM-controlled Diffusion Models},
    booktitle = {Proceedings of the IEEE/CVF Conference on Computer Vision and Pattern Recognition (CVPR)},
    month     = {June},
    year      = {2024},
    pages     = {6327-6336}
}

@article{VSR,
  author    = {Fangyu Liu and Guy Emerson and Nigel Collier},
  title     = {Visual Spatial Reasoning},
  journal   = {Transactions of the Association for Computational Linguistics},
  volume    = {11},
  pages     = {635--651},
  year      = {2023},
  publisher = {MIT Press},
  doi       = {10.1162/tacl_a_00566},
  url       = {https://aclanthology.org/2023.tacl-1.37/}
}

@inproceedings{SpatialVLM,
    author = {Chen, Boyuan and Xu, Zhuo and Kirmani, Sean and Ichter, Brian
              and Sadigh, Dorsa and Guibas, Leonidas and Xia, Fei},
    title = {{SpatialVLM}: Endowing Vision-Language Models with Spatial Reasoning Capabilities},
    booktitle = {Proceedings of the IEEE/CVF Conference on Computer Vision and Pattern Recognition (CVPR)},
    year = {2024},
    pages = {14455--14465}
}

@inproceedings{Spatial457,
    author = {Wang, Xingrui and Ma, Wufei and Zhang, Tiezheng and de Melo, Celso M and Chen, Jieneng and Yuille, Alan},
    title = {Spatial457: A Diagnostic Benchmark for 6D Spatial Reasoning of Large Multimodal Models},
    booktitle = {Proceedings of the IEEE/CVF Conference on Computer Vision and Pattern Recognition (CVPR)},
    year = {2025},
    pages = {24669--24679}
}

@techreport{GPT4o,
  author       = {{OpenAI}},
  title        = {Addendum to GPT‑4o System Card: Native Image Generation},
  institution  = {OpenAI},
  type         = {Technical Report},
  number       = {Native\_Image\_Generation\_System\_Card},
  address      = {San Francisco, CA},
  month        = mar,
  year         = {2025},
  note         = {Available at: \url{https://cdn.openai.com/11998be9-5319-4302-bfbf-1167e093f1fb/Native_Image_Generation_System_Card.pdf}},
}

@misc{blackforestlabs2024flux1dev,
  author = {{Black Forest Labs}},
  title  = {{FLUX.1 [dev]}},
  year   = {2024},
  note   = {Model card},
  url    = {https://huggingface.co/black-forest-labs/FLUX.1-dev}
}

@article{
LMD,
title={{LLM}-grounded Diffusion: Enhancing Prompt Understanding of Text-to-Image Diffusion Models with Large Language Models},
author={Long Lian and Boyi Li and Adam Yala and Trevor Darrell},
journal={Transactions on Machine Learning Research},
issn={2835-8856},
year={2024},
url={https://openreview.net/forum?id=hFALpTb4fR},
note={Featured Certification}
}

@inproceedings{SAM,
    author = {Kirillov, Alexander and Mintun, Eric and Ravi, Nikhila and Mao, Hanzi
              and Rolland, Chloe and Gustafson, Laura and Xiao, Tete
              and Whitehead, Spencer and Berg, Alexander C. and Lo, Wan-Yen
              and Dollár, Piotr and Girshick, Ross},
    title = {Segment Anything},
    booktitle = {Proceedings of the IEEE/CVF International Conference on Computer Vision (ICCV)},
    year = {2023},
    pages = {4015--4026}
}

@inproceedings{ControlNet,
    author = {Zhang, Lvmin and Rao, Anyi and Agrawala, Maneesh},
    title = {Adding Conditional Control to Text-to-Image Diffusion Models},
    booktitle = {Proceedings of the IEEE/CVF International Conference on Computer Vision (ICCV)},
    year = {2023},
    pages = {3836--3847}
}

@inproceedings{Cot-liz,
    title = {CoT-lized Diffusion: Let's Reinforce T2I Generation Step-by-step},
    author = {Liu, Zheyuan and Ning, Munan and Zhang, Qihui and Yang, Shuo
              and Wang, Zhongrui and Yang, Yiwei and Xu, Xianzhe and Song, Yibing
              and Chen, Weihua and Wang, Fan and Yuan, Li},
    booktitle = {Advances in Neural Information Processing Systems},
    volume = {38},
    year = {2025}
}

@inproceedings{DreamSync,
    title = "{D}ream{S}ync: Aligning Text-to-Image Generation with Image Understanding Feedback",
    author = "Sun, Jiao  and
      Fu, Deqing  and
      Hu, Yushi  and
      Wang, Su  and
      Rassin, Royi  and
      Juan, Da-Cheng  and
      Alon, Dana  and
      Herrmann, Charles  and
      Steenkiste, Sjoerd Van  and
      Krishna, Ranjay  and
      Rashtchian, Cyrus",
    editor = "Chiruzzo, Luis  and
      Ritter, Alan  and
      Wang, Lu",
    booktitle = "Proceedings of the 2025 Conference of the Nations of the Americas Chapter of the Association for Computational Linguistics: Human Language Technologies (Volume 1: Long Papers)",
    month = apr,
    year = "2025",
    address = "Albuquerque, New Mexico",
    publisher = "Association for Computational Linguistics",
    url = "https://aclanthology.org/2025.naacl-long.304/",
    doi = "10.18653/v1/2025.naacl-long.304",
    pages = "5920--5945",
    ISBN = "979-8-89176-189-6"
}

@misc{Grape,
        title={{GraPE}: A Generate-Plan-Edit Framework for Compositional T2I Synthesis}, 
        author={Ashish Goswami and Satyam Kumar Modi and Santhosh Rishi Deshineni and Harman Singh and Prathosh A. P and Parag Singla},
        year={2024},
        eprint={2412.06089},
        archivePrefix={arXiv},
        primaryClass={cs.CV},
        url={https://arxiv.org/abs/2412.06089}, 
  }

@article{AttentionMapWang,
  title   = {Compositional Text-to-Image Synthesis with Attention Map Control of Diffusion Models},
  author  = {Wang, Ruichen and Chen, Zekang and Chen, Chen and Ma, Jian and Lu, Haonan and Lin, Xiaodong},
  journal = {Proceedings of the AAAI Conference on Artificial Intelligence},
  volume  = {38},
  number  = {6},
  pages   = {5544--5552},
  year    = {2024},
  doi     = {10.1609/aaai.v38i6.28364},
}

@inproceedings{li2023gligen,
    title = {{GLIGEN}: Open-Set Grounded Text-to-Image Generation},
    author = {Li, Yuheng and Liu, Haotian and Wu, Qingyang and Mu, Fangzhou
              and Yang, Jianwei and Gao, Jianfeng and Li, Chunyuan and Lee, Yong Jae},
    booktitle = {Proceedings of the IEEE/CVF Conference on Computer Vision and Pattern Recognition (CVPR)},
    year = {2023},
    pages = {22511--22521}
}

@article{TENBRINK2011704,
title = {Reference frames of space and time in language},
journal = {Journal of Pragmatics},
volume = {43},
number = {3},
pages = {704-722},
year = {2011},
note = {The Language of Space and Time},
issn = {0378-2166},
doi = {https://doi.org/10.1016/j.pragma.2010.06.020},
url = {https://www.sciencedirect.com/science/article/pii/S037821661000192X},
author = {Thora Tenbrink}
}

@book{Levinson_2003, place={Cambridge}, series={Language Culture and Cognition}, title={Space in Language and Cognition: Explorations in Cognitive Diversity}, publisher={Cambridge University Press}, author={Levinson, Stephen C.}, year={2003}, collection={Language Culture and Cognition}}

@ARTICLE{cueBySeenAndHear,
  
AUTHOR={Coventry, Kenny R.  and Andonova, Elena  and Tenbrink, Thora  and Gudde, Harmen B.  and Engelhardt, Paul E. },
         
TITLE={Cued by What We See and Hear: Spatial Reference Frame Use in Language},
        
JOURNAL={Frontiers in Psychology},
        
VOLUME={Volume 9 - 2018},

YEAR={2018},

URL={https://www.frontiersin.org/journals/psychology/articles/10.3389/fpsyg.2018.01287},

DOI={10.3389/fpsyg.2018.01287},

ISSN={1664-1078}}

@ARTICLE{Mou2002-tp,
  title    = "Intrinsic frames of reference in spatial memory",
  author   = "Mou, Weimin and McNamara, Timothy P",
  journal  = "J Exp Psychol Learn Mem Cogn",
  volume   =  28,
  number   =  1,
  pages    = "162--170",
  month    =  jan,
  year     =  2002,
  address  = "United States",
  language = "en"
}

@InProceedings{SD1-5,
        author    = {Rombach, Robin and Blattmann, Andreas and Lorenz, Dominik and Esser, Patrick and Ommer, Bj\"orn},
        title     = {High-Resolution Image Synthesis With Latent Diffusion Models},
        booktitle = {Proceedings of the IEEE/CVF Conference on Computer Vision and Pattern Recognition (CVPR)},
        month     = {June},
        year      = {2022},
        pages     = {10684-10695}
    }

@inproceedings{
couairon2023diffedit,
title={{DiffEdit}: Diffusion-based semantic image editing with mask guidance},
author={Guillaume Couairon and Jakob Verbeek and Holger Schwenk and Matthieu Cord},
booktitle={The Eleventh International Conference on Learning Representations },
year={2023},
url={https://openreview.net/forum?id=3lge0p5o-M-}
}

@misc{stabilityai2024sd35large,
  title        = {Stable Diffusion 3.5 Large},
  author       = {{Stability AI}},
  year         = {2024},
  howpublished = {\url{https://huggingface.co/stabilityai/stable-diffusion-3.5-large}},
  note         = {Multimodal Diffusion Transformer (MMDiT) text‑to‑image model with 8.1 billion parameters; released under Stability AI Community License},
}

@misc{openai2025_gpt4o_image1,
  author = {{OpenAI}},
  title  = {Introducing Our Latest Image Generation Model in the {API}},
  year   = {2025},
  month  = apr,
  url    = {https://openai.com/index/image-generation-api/},
  note   = {Published April 23, 2025}
}

@misc{Qwen3,
      title={Qwen3 Technical Report}, 
      author={{Qwen Team}},
      year={2025},
      eprint={2505.09388},
      archivePrefix={arXiv},
      primaryClass={cs.CL},
      url={https://arxiv.org/abs/2505.09388}, 
}

@inproceedings{OWLv2,
    title = {Scaling Open-Vocabulary Object Detection},
    author = {Minderer, Matthias and Gritsenko, Alexey A. and Houlsby, Neil},
    booktitle = {Advances in Neural Information Processing Systems},
    year = {2023}
}

@inproceedings{DPT,
    author = {Ranftl, René and Bochkovskiy, Alexey and Koltun, Vladlen},
    title = {Vision Transformers for Dense Prediction},
    booktitle = {Proceedings of the IEEE/CVF International Conference on Computer Vision (ICCV)},
    year = {2021},
    pages = {12179--12188},
    doi = {10.1109/ICCV48922.2021.01196}
}

@inproceedings{orient_anything,
    title = {Orient Anything: Learning Robust Object Orientation Estimation from Rendering {3D} Models},
    author = {Wang, Zehan and Zhang, Ziang and Pang, Tianyu and Du, Chao
              and Zhao, Hengshuang and Zhao, Zhou},
    booktitle = {Proceedings of the 42nd International Conference on Machine Learning},
    series = {Proceedings of Machine Learning Research},
    volume = {267},
    pages = {65556--65574},
    year = {2025}
}

@inproceedings{mirzaee-etal-2021-spartqa,
    title = "{SPARTQA}: A Textual Question Answering Benchmark for Spatial Reasoning",
    author = "Mirzaee, Roshanak  and
      Rajaby Faghihi, Hossein  and
      Ning, Qiang  and
      Kordjamshidi, Parisa",
    editor = "Toutanova, Kristina  and
      Rumshisky, Anna  and
      Zettlemoyer, Luke  and
      Hakkani-Tur, Dilek  and
      Beltagy, Iz  and
      Bethard, Steven  and
      Cotterell, Ryan  and
      Chakraborty, Tanmoy  and
      Zhou, Yichao",
    booktitle = "Proceedings of the 2021 Conference of the North American Chapter of the Association for Computational Linguistics: Human Language Technologies",
    month = jun,
    year = "2021",
    address = "Online",
    publisher = "Association for Computational Linguistics",
    url = "https://aclanthology.org/2021.naacl-main.364/",
    doi = "10.18653/v1/2021.naacl-main.364",
    pages = "4582--4598"
}

@inproceedings{mirzaee-kordjamshidi-2022-transfer,
    title = "Transfer Learning with Synthetic Corpora for Spatial Role Labeling and Reasoning",
    author = "Mirzaee, Roshanak  and
      Kordjamshidi, Parisa",
    editor = "Goldberg, Yoav  and
      Kozareva, Zornitsa  and
      Zhang, Yue",
    booktitle = "Proceedings of the 2022 Conference on Empirical Methods in Natural Language Processing",
    month = dec,
    year = "2022",
    address = "Abu Dhabi, United Arab Emirates",
    publisher = "Association for Computational Linguistics",
    url = "https://aclanthology.org/2022.emnlp-main.413/",
    doi = "10.18653/v1/2022.emnlp-main.413",
    pages = "6148--6165"
}

@inproceedings{shi2022stepgamenewbenchmarkrobust,
title={StepGame: A New Benchmark for Robust Multi-Hop Spatial Reasoning in Texts},
author={Shi, Zhengxiang and Zhang, Qiang and Lipani, Aldo},
volume={36},
url={https://ojs.aaai.org/index.php/AAAI/article/view/21383},
DOI={10.1609/aaai.v36i10.21383}, 
booktitle={Proceedings of the AAAI Conference on Artificial Intelligence},
year={2022},
month={Jun.},
pages={11321-11329}
}

@inproceedings{
zhang2025spartund,
title={{SPARTUN}{3D}: Situated Spatial Understanding of {3D} World in Large Language Model},
author={Yue Zhang and Zhiyang Xu and Ying Shen and Parisa Kordjamshidi and Lifu Huang},
booktitle={The Thirteenth International Conference on Learning Representations},
year={2025},
url={https://openreview.net/forum?id=FGMkSL8NR0}
}

@inproceedings{vpeval,
    title = {Visual Programming for Step-by-Step Text-to-Image Generation and Evaluation},
    author = {Cho, Jaemin and Zala, Abhay and Bansal, Mohit},
    booktitle = {Advances in Neural Information Processing Systems},
    volume = {36},
    pages = {6048--6069},
    year = {2023}
}

@inproceedings{mattersim,
  title={Vision-and-Language Navigation: Interpreting visually-grounded navigation instructions in real environments},
  author={Peter Anderson and Qi Wu and Damien Teney and Jake Bruce and Mark Johnson and Niko S{\"u}nderhauf and Ian Reid and Stephen Gould and Anton van den Hengel},
  booktitle={Proceedings of the IEEE Conference on Computer Vision and Pattern Recognition (CVPR)},
  year={2018}
}

@InProceedings{FreeControl_2024,
    author    = {Mo, Sicheng and Mu, Fangzhou and Lin, Kuan Heng and Liu, Yanli and Guan, Bochen and Li, Yin and Zhou, Bolei},
    title     = {FreeControl: Training-Free Spatial Control of Any Text-to-Image Diffusion Model with Any Condition},
    booktitle = {Proceedings of the IEEE/CVF Conference on Computer Vision and Pattern Recognition (CVPR)},
    month     = {June},
    year      = {2024},
    pages     = {7465-7475}
}

@inproceedings{
pang2024attndreambooth,
title={AttnDreamBooth: Towards Text-Aligned Personalized Text-to-Image Generation},
author={Lianyu Pang and Jian Yin and Baoquan Zhao and Feize Wu and Fu Lee Wang and Qing Li and Xudong Mao},
booktitle={The Thirty-eighth Annual Conference on Neural Information Processing Systems},
year={2024},
url={https://openreview.net/forum?id=4bINoegDcm}
}

@article{huang2023t2icompbench,
  title={T2i-compbench: A comprehensive benchmark for open-world compositional text-to-image generation},
  author={Huang, Kaiyi and Sun, Kaiyue and Xie, Enze and Li, Zhenguo and Liu, Xihui},
  journal={Advances in Neural Information Processing Systems},
  volume={36},
  pages={78723--78747},
  year={2023}
}

@inproceedings{DiffusionModel,
  title     = {Denoising Diffusion Probabilistic Models},
  author    = {Ho, Jonathan and Jain, Ajay and Abbeel, Pieter},
  booktitle = {Advances in Neural Information Processing Systems (NeurIPS)},
  year      = {2020},
  volume    = {33},
  pages     = {6840--6851},
  url       = {https://proceedings.neurips.cc/paper/2020/file/4c5bcfec8584af0d967f1ab10179ca4b-Paper.pdf}
}

@misc{midas3-1,
      title={{MiDaS} v3.1 -- A Model Zoo for Robust Monocular Relative Depth Estimation}, 
      author={Reiner Birkl and Diana Wofk and Matthias Müller},
      year={2023},
      eprint={2307.14460},
      archivePrefix={arXiv},
      primaryClass={cs.CV},
      url={https://arxiv.org/abs/2307.14460}, 
}

@inproceedings{grounding_dino,
    title = {{Grounding DINO}: Marrying DINO with Grounded Pre-Training for Open-Set Object Detection},
    author = {Liu, Shilong and Zeng, Zhaoyang and Ren, Tianhe and Li, Feng
              and Zhang, Hao and Yang, Jie and Jiang, Qing and Li, Chunyuan
              and Yang, Jianwei and Su, Hang and Zhu, Jun and Zhang, Lei},
    booktitle = {Proceedings of the European Conference on Computer Vision (ECCV)},
    year = {2024},
    doi = {10.1007/978-3-031-72970-6_3}
}

@inproceedings{kordjamshidi-etal-2010-spatial,
    title = "Spatial Role Labeling: Task Definition and Annotation Scheme",
    author = "Kordjamshidi, Parisa  and
      Van Otterlo, Martijn  and
      Moens, Marie-Francine",
    editor = "Calzolari, Nicoletta  and
      Choukri, Khalid  and
      Maegaard, Bente  and
      Mariani, Joseph  and
      Odijk, Jan  and
      Piperidis, Stelios  and
      Rosner, Mike  and
      Tapias, Daniel",
    booktitle = "Proceedings of the Seventh International Conference on Language Resources and Evaluation ({LREC}'10)",
    month = may,
    year = "2010",
    address = "Valletta, Malta",
    publisher = "European Language Resources Association (ELRA)",
    url = "https://aclanthology.org/L10-1584/"
}

@inproceedings{cognitive_map,
    title = {Thinking in Space: How Multimodal Large Language Models See, Remember, and Recall Spaces},
    author = {Yang, Jihan and Yang, Shusheng and Gupta, Anjali W. and Han, Rilyn
              and Fei-Fei, Li and Xie, Saining},
    booktitle = {Proceedings of the IEEE/CVF Conference on Computer Vision and Pattern Recognition (CVPR)},
    year = {2025},
    pages = {10632--10643},
    doi = {10.1109/CVPR52734.2025.00994}
}

@misc{gemini3proimage2025,
  author = {{Google DeepMind}},
  title  = {{Gemini 3 Pro Image} Model Card},
  year   = {2025},
  month  = nov,
  url    = {https://deepmind.google/models/model-cards/gemini-3-pro-image/}
}

@misc{wu2025qwenimagetechnicalreport,
      title={{Qwen-Image} Technical Report}, 
      author={Chenfei Wu and Jiahao Li and Jingren Zhou and Junyang Lin and Kaiyuan Gao and Kun Yan and Sheng-ming Yin and Shuai Bai and Xiao Xu and Yilei Chen and Yuxiang Chen and Zecheng Tang and Zekai Zhang and Zhengyi Wang and An Yang and Bowen Yu and Chen Cheng and Dayiheng Liu and Deqing Li and Hang Zhang and Hao Meng and Hu Wei and Jingyuan Ni and Kai Chen and Kuan Cao and Liang Peng and Lin Qu and Minggang Wu and Peng Wang and Shuting Yu and Tingkun Wen and Wensen Feng and Xiaoxiao Xu and Yi Wang and Yichang Zhang and Yongqiang Zhu and Yujia Wu and Yuxuan Cai and Zenan Liu},
      year={2025},
      eprint={2508.02324},
      archivePrefix={arXiv},
      primaryClass={cs.CV},
      url={https://arxiv.org/abs/2508.02324}, 
}

@techreport{openai2025o3o4mini,
  author      = {{OpenAI}},
  title       = {{OpenAI o3 and o4-mini System Card}},
  institution = {OpenAI},
  year        = {2025},
  month       = apr,
  url         = {https://openai.com/index/o3-o4-mini-system-card/}
}

@misc{flux1fill,
  author = {{Black Forest Labs}},
  title  = {{FLUX.1 Fill [dev]}},
  year   = {2024},
  note   = {Model card},
  url    = {https://huggingface.co/black-forest-labs/FLUX.1-Fill-dev}
}
